\documentclass[conference]{IEEEtran}
\pdfoutput=1    % For arXiv issues

\usepackage{xcolor}
\definecolor{wong-black}        {HTML}{000000}
\definecolor{wong-lightorange}  {HTML}{E69F00}
\definecolor{wong-lightblue}    {HTML}{56B4E9}
\definecolor{wong-green}        {HTML}{009E73}
\definecolor{wong-yellow}       {HTML}{F0E442}
\definecolor{wong-darkblue}     {HTML}{0072B2}
\definecolor{wong-darkorange}   {HTML}{D55E00}
\definecolor{wong-pink}         {HTML}{CC79A7}

\usepackage[accsupp]{axessibility}  % Improves PDF readability for those with disabilities.

\usepackage{url}
\usepackage{hyperref} % Working hyperlink (https://www.overleaf.com/learn/latex/Hyperlinks)

\hypersetup{
    colorlinks=true,
    citecolor=wong-green,
    linkcolor=wong-darkblue,
    filecolor=wong-pink,      
    urlcolor=wong-darkblue,
    pdfpagemode=FullScreen,
    }

\usepackage[english]{babel}

\usepackage{cite}
\usepackage{amsmath,amssymb,amsfonts}
\usepackage{algorithmic}
\usepackage{graphicx}
\usepackage{booktabs}
\usepackage{textcomp}
\usepackage[nolist, nohyperlinks, printonlyused]{acronym} % For consistent acronyms

\usepackage{lscape}
\usepackage{subfig}

\def\BibTeX{{\rm B\kern-.05em{\sc i\kern-.025em b}\kern-.08em
    T\kern-.1667em\lower.7ex\hbox{E}\kern-.125emX}}

\usepackage[normalem]{ulem}

\usepackage{pifont}% http://ctan.org/pkg/pifont
\newcommand{\cmark}{\ding{51}}%
\newcommand{\xmark}{\ding{55}}%

\begin{document}
\bstctlcite{IEEEexample:BSTcontrol}

% -------------------------- TITLE -------------------------------

\title{Perception Datasets for Anomaly Detection in\\Autonomous Driving: A Survey}

% -------------------------- AUTHORS -------------------------------

\author{\IEEEauthorblockN{Daniel Bogdoll\IEEEauthorrefmark{1}\IEEEauthorrefmark{2},
Svenja Uhlemeyer\IEEEauthorrefmark{3}\IEEEauthorrefmark{4},
Kamil Kowol\IEEEauthorrefmark{3}\IEEEauthorrefmark{4},
J. Marius Zöllner\IEEEauthorrefmark{1}\IEEEauthorrefmark{2}}

\IEEEauthorblockA{\IEEEauthorrefmark{1}FZI Research Center for Information Technology, Germany}
\IEEEauthorblockA{\IEEEauthorrefmark{2}Karlsruhe Institute of Technology, Germany\\}
\IEEEauthorblockA{\IEEEauthorrefmark{3}University of Wuppertal, Germany\\}
\IEEEauthorblockA{\IEEEauthorrefmark{4}Interdisciplinary Center for Machine Learning and Data Analytics, Germany\\}
}

\maketitle

% -------------------------- ACRONYMS -------------------------------

\begin{acronym}
    \acro{ml}[ML]{Machine Learning}
	\acro{cnn}[CNN]{Convolutional Neural Network}
	\acro{dl}[DL]{Deep Learning}
	\acro{ad}[AD]{Autonomous Driving}
\end{acronym}

% -------------------------- ABSTRACT -------------------------------

\begin{abstract}
Deep neural networks (DNN) which are employed in perception systems for autonomous driving require a huge amount of data to train on, as they must reliably achieve high performance in all kinds of situations. However, these DNN are usually restricted to a closed set of semantic classes available in their training data, and are therefore unreliable when confronted with previously unseen instances. Thus, multiple perception datasets have been created for the evaluation of anomaly detection methods, which can be categorized into three groups: real anomalies in real-world, synthetic anomalies augmented into real-world and completely synthetic scenes.
This survey provides a structured and, to the best of our knowledge, complete overview and comparison of perception datasets for anomaly detection in autonomous driving. Each chapter provides information about tasks and ground truth, context information, and licenses.
Additionally, we discuss current weaknesses and gaps in existing datasets to underline the importance of developing further data.
\end{abstract}

% -------------------------- KEYWORDS -------------------------------


% -------------------------- CONTENT -------------------------------

\section{Introduction}
\label{sec:introduction}

When thinking about autonomous vehicles that move safely through traffic, it is necessary to perceive the environment correctly in order to provide safe driving. To ensure this, DNN must be extensively trained and tested with data required to solve the task. In this context, numerous datasets have been created for use in road traffic~\cite{Guo2020OverviewDatasets,Liu2021OverviewDatasets,Bogdoll_Addatasets_2022_VEHITS}, most of which include daytime and sunny weather and harmless everyday scenes. As numerous new datasets are published each year~\cite{Bogdoll_Impact_2023_arxiv}, it is important to include anomalies, out-of-distribution (OOD) instances, novelties, outlier, and corner cases, which primarily describe what is unknown or unusual~\cite{Breitenstein_Systematization_2020_IV,Heidecker_Application_2021_IV,Bogdoll_Description_2022_ICCV}, to improve the detection and eventually handling of safety-critical driving situations. To deal with such circumstances, the field of anomaly detection is a highly active research field~\cite{Bogdoll_Anomaly_2022_CVPR, du2022vos, Bogdoll_Multimodal_2022_SMC, Uhlemeyer_Towards_2022_UAI, Bogdoll_Experiments_2022_ICECCME,  Chan_Detecting_2022_Springer}. However, most public datasets follow a closed world assumption~\cite{Reiter_Closed_1978_Springer} and offer no room to detect anomalies.\\

In this work, to the best of our knowledge, we offer a complete collection of perception datasets with labeled anomalies in the domain of autonomous driving. These datasets show a strong focus on object- and scene-level anomalies, as described by Breitenstein et al.~\cite{Breitenstein_Systematization_2020_IV}. We visualize the number and distribution of anomalies and provide insights and research gaps for future work. We have included datasets, which 
\begin{itemize}
    \item are public and available, as of 01 February 2023
    \item provide sensor data from the ego-perspective, given that licenses of potentially utilized datasets allow for that
    \item include pixel- or point-wise anomaly labels, at least in the form of a small validation set
\end{itemize}

Works, which we have excluded, include SiMOOD~\cite{senaferreira:hal-03779723}, as they only provide a framework, but no raw data; TOR4D, Rare4D~\cite{Wong_Identifying_2019_CORL} and FS~Web~\cite{Blum_Fishyscapes_2019_ICCV}, as they are not public; MUAD~\cite{franchi2022muad}, \mbox{DANGER-vKITTI}, \mbox{DANGER-vKITTI2}~\cite{Xu_DANGER_2022_BayLearn, Xu_Framework_2022_Neurips}, and \mbox{FDP-set}~\cite{LEE2023119242}, as they are not yet published; and WOS~\cite{Maag_Video_2022_ACCV}, as there are no anomaly labels provided. We also excluded works which focus on adverse conditions, such as WildDash~\cite{Zendel_2018_ECCV, Zendel_2022_CVPR}, ACDC~\cite{sakaridisACDCAdverseConditions2021} or Rain Augmentation~\cite{tremblayRainRenderingEvaluating2021}, as they classify entire scenes, which does not allow inferring the potential relevance of anomalies.

Next to perception datasets, there are also trajectory datasets or frameworks which include anomalies~\cite{Roesch_Space_2022_SSCI}, such as \mbox{R-U-MAAD}~\cite{wiederer_benchmark_2022}, KING~\cite{hanselmann_king_2022} or STRIVE~\cite{Vadis_generating_2022_CVPR}. While such anomalies are among the most challenging ones, those approaches do not provide sensory perception data, but can only be executed in simulation, which does not provide unambiguous visual environment representations.

\textbf{Research Gap.}
While there are numerous new datasets in the field of autonomous driving published each year~\cite{Bogdoll_Impact_2023_arxiv}, there are only very few works that focus on dataset analysis in general and even less that focus on the field of anomaly detection. While there exists a recent overview of anomaly detection methods~\cite{Bogdoll_Anomaly_2022_CVPR}, there is a lack of structured knowledge related to datasets containing anomalies, although such anomalies or corner cases are currently core limiting factors for scaling autonomous vehicles.

\textbf{Contribution.}
Our work aims to aid researchers in the field of anomaly detection to gain an overview of all relevant datasets that include anomalies. We provide clear selection criteria and point to specifically excluded datasets for an even broader horizon. We provide detailed, structured information and visualizations on 16 datasets in their historical order in Section~\ref{sec:relatedwork}. Our survey is the only one of its kind that provides a detailed overview of currently available perception datasets for anomaly detection. In Section~\ref{sec:conclusion}, we provide an extensive discussion on similarities, issues, and research gaps. All code to recreate our visualizations is available on \href{https://github.com/daniel-bogdoll/anomaly_datasets}{GitHub}.

\begin{table*}[t]
\resizebox{\textwidth}{!}{%
\begin{tabular}{@{}lclcclcccr@{}}
\toprule
\textbf{Dataset}                & \textbf{Year} & \textbf{Sensors} & \textbf{Size (Test/Val)}                                              & \textbf{Resolution}                                                             & \textbf{Anomaly Source} & \textbf{Temporal}     & \textbf{\#OOD Classes} & \textbf{Groundtruth} \\ \midrule
\textbf{Fishyscapes}~\cite{Blum_Fishyscapes_2019_ICCV,Blum_Fishyscapes_2021_IJCV,fishyscapes_web}            &               &                  &                                                                       &                                                                                &                                               &                       &                    &                      \\
FS Lost and Found       & 2019          & Camera           & 275 / 100                                                             & $2048 \times 1024$                                                             &                     Recording               & \xmark &          1          &       Semantic Mask               \\
FS Static              & 2019          & Camera           & 1,000 / 30                                                             & $2048 \times 1024$                                                             &                     Data Augmentation          &         \xmark              &     1               &        Semantic Mask              \\ \midrule
\textbf{CAOS}~\cite{Hendrycks_Scaling_2022_ICML, caos_web}                   &               &                  &                                                                       &                                                                                &                    &                           &                       &                    &                      \\
StreetHazards                   & 2019          & Camera           &        1,500                                                               & $1280 \times 720$                                                              &                     Simulation             &            \cmark           &        1            &      Semantic Mask                \\
BDD-Anomaly                     & 2019          & Camera           &     810                                                                  & $1280 \times 720$                                                              &                     Class Exclusion       &            \xmark           &           3         &        Semantic Mask              \\ \midrule
\textbf{SegmentMeIfYouCan}~\cite{Chan_SegmentMeIfYouCan_2021_NEURIPS, smiyc_web}      &               &                  &                                                                       &                                                                                &                    &                           &                       &                    &                      \\
RoadAnomaly21                   & 2021          & Camera           & 100 / 10                                                              & \begin{tabular}[c]{@{}l@{}}$2048 \times 1024$\\ $1280 \times 720$\end{tabular} &                     Web Sourcing               & \xmark                &          1          &           Semantic Mask           \\
RoadObstacle21                  & 2021          & Camera           & \begin{tabular}[c]{@{}c@{}}327~(+55) / 30\end{tabular}    & $1920 \times 1080$                                                             &                     Recording               & \cmark             &          1          &      Semantic Mask                \\ \midrule
\textbf{CODA}~\cite{Li_CODA_2022_ECCV, coda_web}                   &               &                  &                                                                       &                                                                                                    &                           &                       &                    &                      \\
CODA-KITTI                 & 2022          & Camera, Lidar    &     309                                                                  &        $1242 \times 1376$                                                                        &                     Void Classes               &           \xmark            &             6      & Bounding Boxes       \\
CODA-nuScenes              & 2022          & Camera, Lidar    &   134                                                                    &             $1600 \times 900$                                                                   &                     Void Classes              &           \xmark            &         17           &      Bounding Boxes                \\
CODA-ONCE                  & 2022          & Camera, Lidar    &    1,057                                                                   &        $1920 \times 1020$                                                                        &                     Automated OOD Proposal               &           \xmark            &    32                &     Bounding Boxes                 \\
CODA2022-ONCE                & 2022          & Camera, Lidar           &  717                                                                     &     $1355 \times 720$                                                                           &                     Automated OOD Proposal               &       \xmark                &     29               &         Bounding Boxes             \\
CODA2022-SODA10M                   & 2022          & Camera           &    4,167                                                                   &        \begin{tabular}[c]{@{}c@{}}$1280 \times 720$\\ $958 \times 720$\end{tabular}                                                                        &                     Automated OOD Proposal               &          \xmark             &         29           &       Bounding Boxes               \\ \midrule
\textbf{Wuppertal OOD Tracking}~\cite{Maag_Video_2022_ACCV, wuppertal_web, OOD_Tracking_2022_Git} &               &                  &                                                                       &                                                                                &                    &                           &                       &                    &                      \\
Street Obstacle Sequences (SOS) & 2022          & Camera, Depth    & \begin{tabular}[c]{@{}l@{}}1,129\end{tabular} & $1920 \times 1080$                                                             &                     Recording               & \cmark              & 13                 & Instance Mask        \\
CARLA-WildLife (CWL)            & 2022          & Camera, Depth    & 1,210                                                                  & $1920 \times 1080$                                                             &                     Simulation            & \cmark              & 18                 &      Instance Mask                \\ \midrule
\textbf{Misc}                   &               &                  &                                                                       &                                                                                &                    &                           &                       &                    &                      \\
Lost and Found~\cite{Pinggera_Lost_2016_IROS,LostandFound_Web}                  & 2016          & Stereo Cameras           & 2,104                                                                  & $2048 \times 1024$                                                             &                     Recording               &  \cmark                     &       42             &                 Semantic Mask     \\
WD-Pascal~\cite{Bevandic_Simultaneous_2019_GCPR, wdpascal_web}                       & 2019          & Camera           &             70                                                          &      $1920 \times 1080$                                                                          &                     Data Augmentation          &   \xmark                    &          1          &               Semantic Mask       \\
Vistas-NP~\cite{visapp21, vistasnp_web}                      & 2020          & Camera           &      11,167                                                                 &               Varying                                                                &                     Class Exclusion        &               \xmark        &            4        &                 Semantic Mask     \\ \bottomrule
\end{tabular}%
}
\caption{Overview over all analyzed datasets, clustered by the benchmark in which they were presented.}
\label{tab:overview}
\end{table*}

\section{Datasets}
\label{sec:relatedwork}

In autonomous driving, the detection of atypical and dangerous situations is crucial for the safety of all road users. In order to improve the ability of today's models to handle such critical situations, datasets are required that allow for targeted training and, more importantly, testing with such critical situations. Therefore, various datasets have emerged in recent years, which we describe in this section. As shown in Table~\ref{tab:overview}, we cluster datasets by their benchmark and categorize them by their \textit{Anomaly Source}:

\textbf{Automated OOD Proposal.} This approach allows for the utilization of large, unlabeled datasets. Here, an automated proposal method is used to generate first anomaly proposals. This can be done with any anomaly detection approach, e.g., uncertainty, intermediate detections, geometric priors, or model contradictions. Subsequently, human experts take care of false positives and refine the proposals.

\textbf{Misc Classes.} Based on a labeled dataset, all regions which are either labeled with \textit{void} or \textit{misc} can be examined further. These terms are often used interchangeably and mostly refer to uncommon objects or irrelevant areas. Human experts then relabel those classes as anomalies, if appropriate.

\textbf{Class Exclusion.} This approach is based on a labeled dataset. Hypothetical anomalies are created by excluding frames with known classes from the train and validation splits. A novel test split is created with these, treating the selected classes as anomalies.

\textbf{Web Sourcing}. In this approach, human experts actively search for images that include atypical classes. As a reference list for known classes, often Cityscapes classes~\cite{Cordts2016TheCD} are used.

\textbf{Recording and Simulation.} Here, anomalies are recorded through data collection by driving in the real world~\cite{Maag_Video_2022_ACCV,LostandFound_Web} or in the synthetic world~\cite{Hendrycks_Scaling_2022_ICML,Kowol_AEye_2022_CHIRA,Bogdoll_Ontology_2022_ECCV}. Often, anomalies are also not included in the Cityscapes classes.

\textbf{Data Augmentation.} For this technique, any dataset can be used as a baseline. By synthetic manipulation of scenes~\cite{Koduri_Aureate_2018_WCX,9827351}, anomalies are pasted onto the original image and can be labeled accordingly. As previously, anomalies are typically not included in the Cityscapes classes.

In addition to distinguishing between different anomaly methods, we also differentiate whether datasets provide single frames with anomalies or scenarios~\cite{ulbrich} with a temporal context. In Figure~\ref{fig:anomalymasks}, we show cumulated anomaly masks for all datasets. These give an overview of the number of included anomalies and in which regions of the images they can be found. In the following chapters, we will introduce each dataset and provide details on the anomalies, possible tasks, the general context, and license agreements. Furthermore, we show examples for each dataset, where \textit{anomaly} and \textit{void} instances are overlaid in orange and black, and outlined in green and red respectively.

\subsection{Lost and Found} % 2016

 %Hard facts: Jahr, Autoren, mit welchen anderen Datensätzen hängt er zusammen?
The Lost and Found dataset~\cite{Pinggera_Lost_2016_IROS} was introduced in 2016 by Pinggera et al., being the first dataset with a focus on the detection of small road hazards, as shown in Figure~\ref{fig:laf}.

\subsubsection{Tasks and Ground truth}
The provided stereo masks for the task of semantic segmentation allow for pixel- and instance-level evaluation, as proposed by the authors. Their instance-level approach is based on a 3D stixel representation, which is very method-specific. As the dataset provides data from stereo cameras, geometric methods can be applied. The anomalies include 42 individual object types that can realistically be found in a street environment. The objects are categorized into \textit{standard objects}, \textit{random hazards}, \textit{emotional hazards} as animals or toys, \textit{random non-hazards}, and \textit{humans} and include both static and dynamic obstacles.

\subsubsection{Context}
The data was collected in the greater Stuttgart area, Germany. It includes irregular road surfaces, large object distances, and illumination changes~\cite{Pinggera_Lost_2016_IROS}. Typical environments include housing areas, parking lots, or industrial areas~\cite{Blum_Fishyscapes_2019_ICCV}.

\subsubsection{License}
The dataset is “freely available to academic and non-academic entities for non-commercial purposes”~\cite{LostandFound_Web}.

\subsection{Fishyscapes} % 05/2019
The Fishyscapes (FS) benchmark~\cite{Blum_Fishyscapes_2021_IJCV} was introduced in 2019 by Blum et al. for the evaluation of anomaly detection methods in semantic segmentation. While most of the data is withheld for evaluation, the authors provide validation sets for the different datasets FS Lost and Found and FS Static. A third FS Web dataset is completely withheld. The first dataset is a subset of the Lost and Found dataset~\cite{Pinggera_Lost_2016_IROS}. The others are based on the Cityscapes~\cite{Cordts2016TheCD} validation data, overlayed with anomalous objects which are either extracted from the generic Pascal VOC~\cite{Everingham2009ThePV} dataset or crawled from the internet. The FS Static validation frames are automatically generated from the Cityscapes dataset.

\begin{figure}[t]
    \captionsetup[subfloat]{labelformat=empty}
    \centering
    \subfloat{\includegraphics[width=0.485\columnwidth]{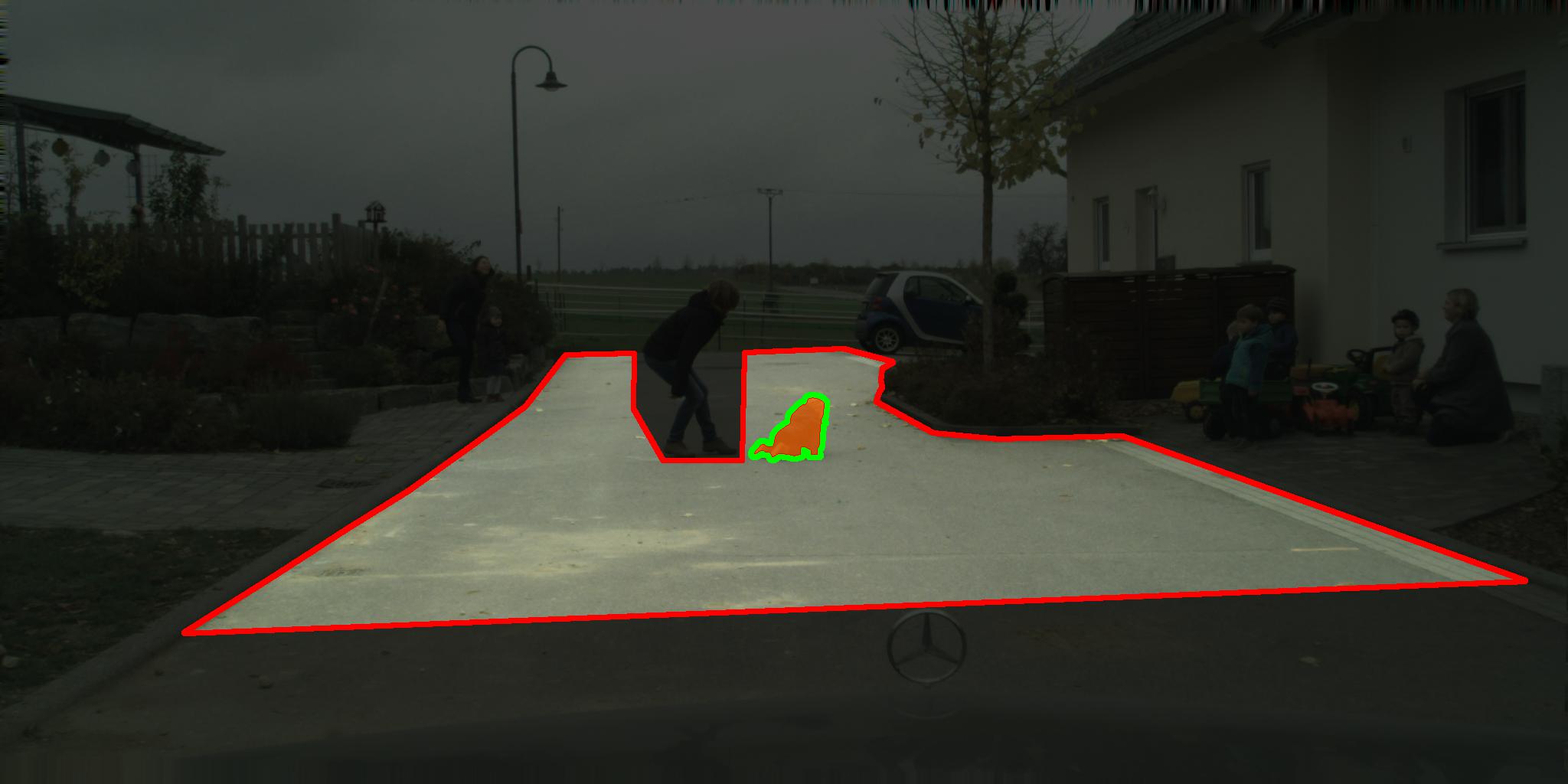}}\hfill
    \subfloat{\includegraphics[width=0.485\columnwidth]{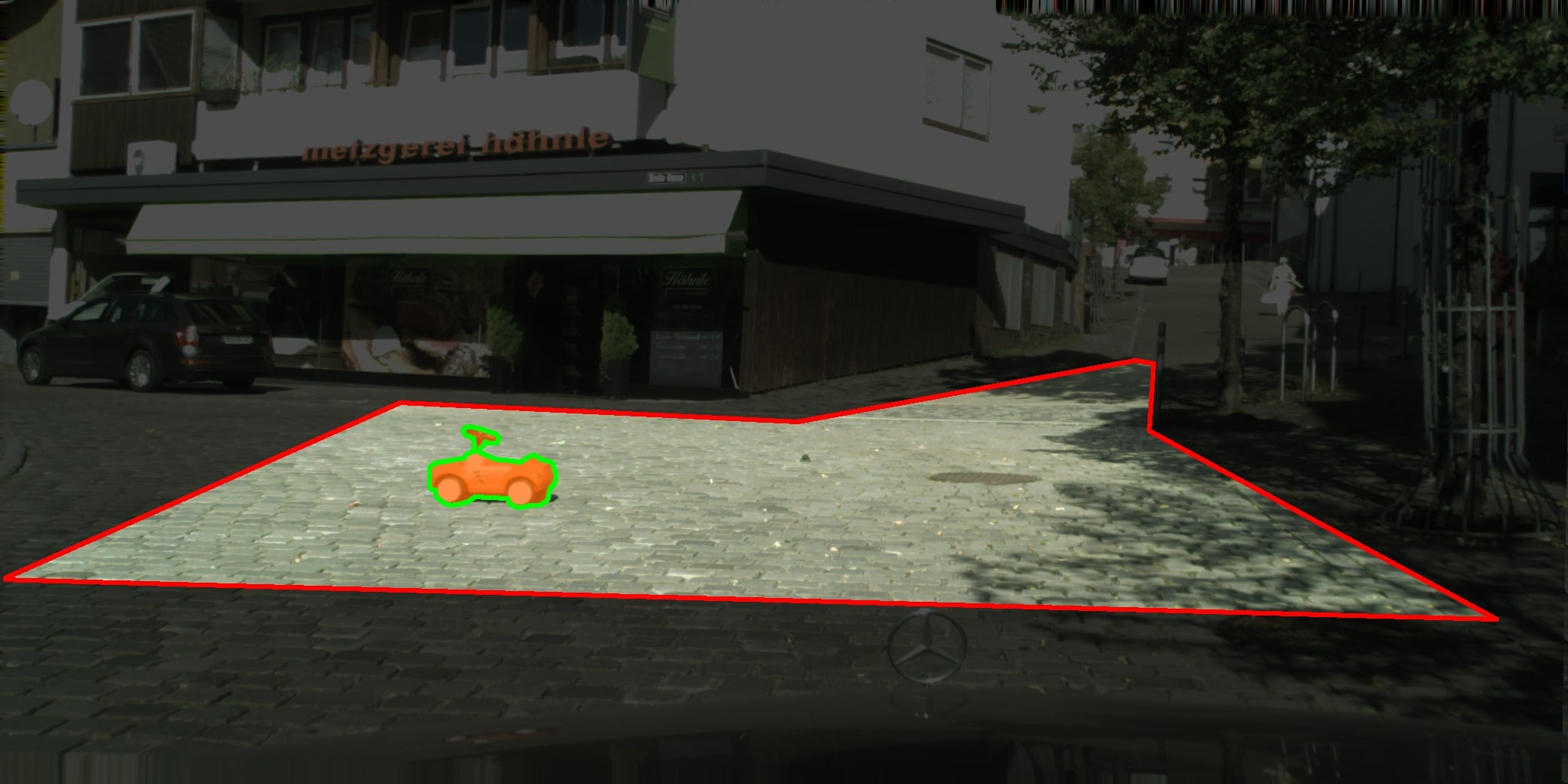}}
    \caption{Lost and Found: Visualization of exemplary real anomaly types in real-world scenes.}
    \label{fig:laf}
\end{figure}

\subsubsection{Tasks and Ground truth}
\begin{figure}[t]
    \captionsetup[subfloat]{labelformat=empty}
    \centering
    \subfloat[FS Lost and Found]{\includegraphics[width=0.485\columnwidth]{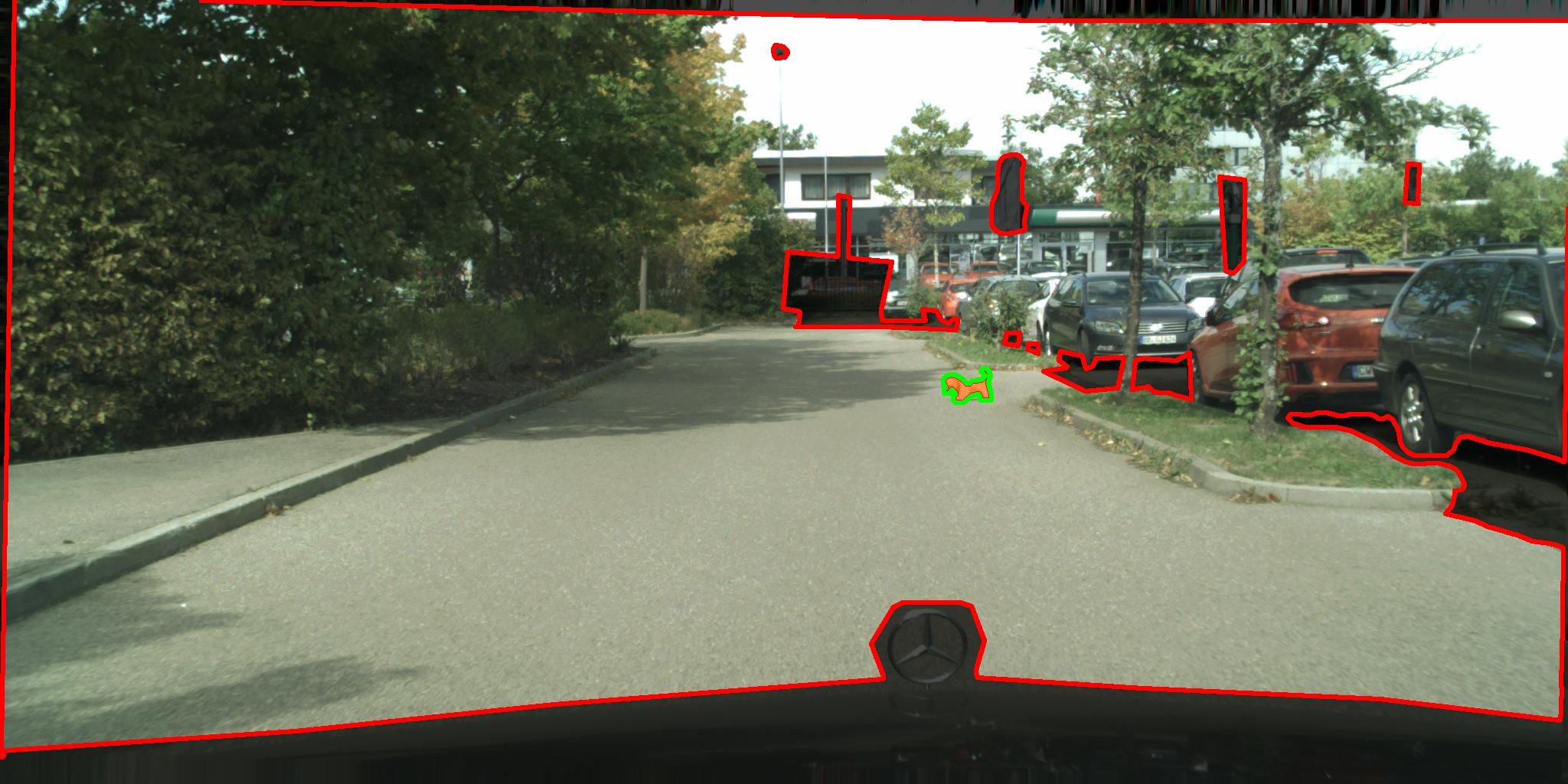}}\hfill
    \subfloat[FS Static]{\includegraphics[width=0.485\columnwidth]{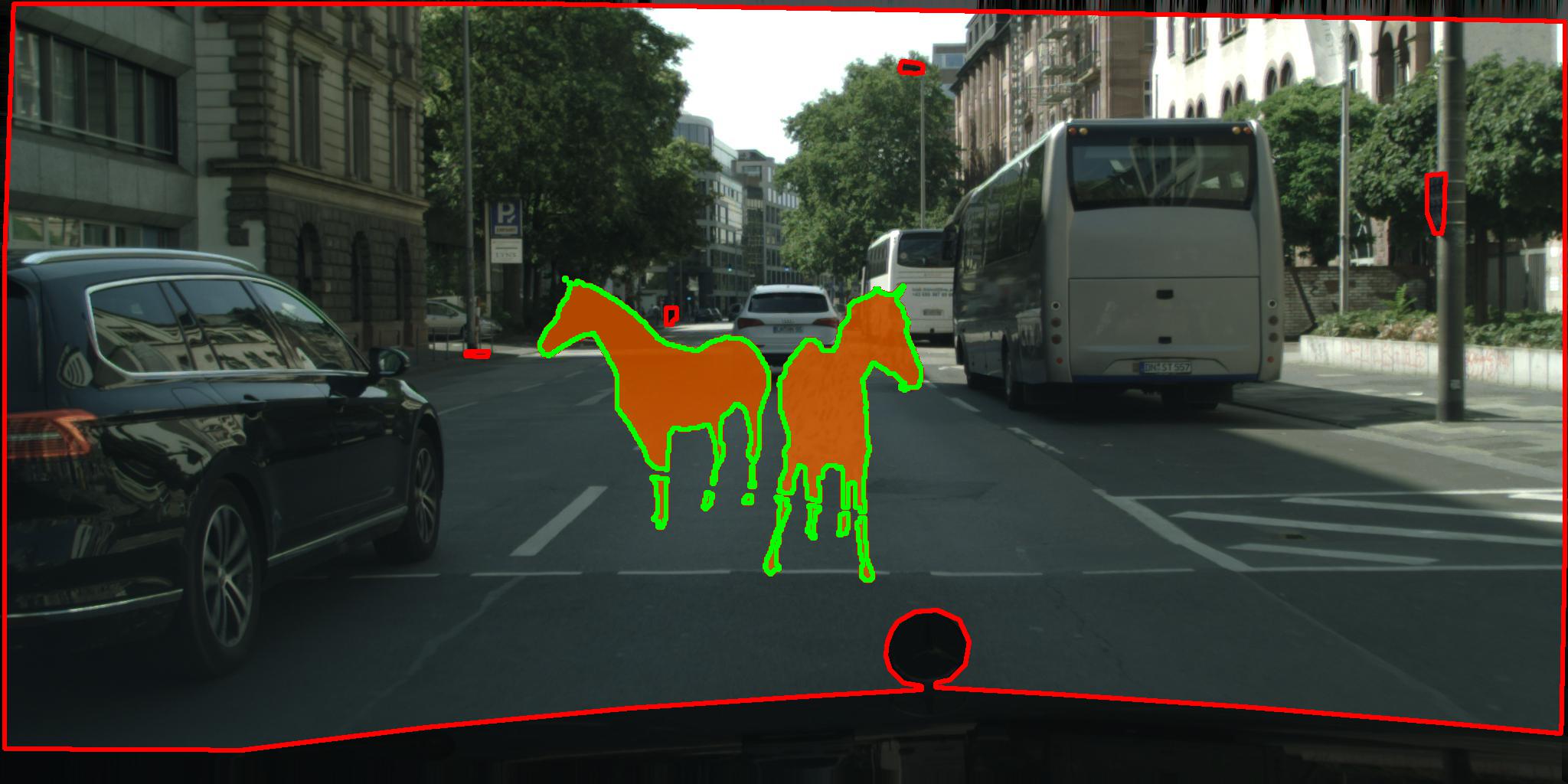}}
    \caption{Fishyscapes: Samples from the val splits, showing real-world scenes with real (left) and synthetic (right) anomalies.}
    \label{fig:fs}
\end{figure}

The FS datasets are designed for the task of semantic segmentation. FS Lost and Found is enriched with fine-grained binary semantic masks. The refined annotation of the background is shown by comparing Figure~\ref{fig:fs} with Figure~\ref{fig:laf}. Furthermore, sequences, where the anomalous objects are bicycles or children, are filtered out, as they can be assigned to one of the Cityscapes classes.
For FS Static and FS Web, novel objects are blended into already annotated scenes from Cityscapes, resulting in fully annotated semantic masks. The anomalies extracted from the Pascal VOC dataset belong to the classes \emph{airplane, bird, boat, bottle, cat, chair, cow, dog, horse, sheep, sofa,} and \emph{tvmonitor}.

\subsubsection{Context}
As all images originate from the Cityscapes or the Lost and Found datasets, they are recorded at daytime under clear weather conditions. For the augmented Cityscapes data, this also entails that the street scenes include instances from known classes, such as humans or other vehicles. 
Depending on the anomaly type, the anomalies have a higher probability to appear either on the lower- or the upper half.

\subsubsection{License}
FS is licensed under the Apache License 2.0.

\subsection{CAOS} % 08/2019
The Combined Anomalous Object Segmentation (CAOS) benchmark~\cite{Hendrycks_Scaling_2022_ICML} was first introduced in 2019 by Hendrycks et al. and includes the datasets StreetHazards and BDD-Anomaly.

StreetHazards is based on the CARLA simulation environment~\cite{Dosovitskiy2017CARLAAO}. The training set includes three towns from CARLA. A further town is reserved for the validation set, and two additional ones are exclusive to the test set. BDD-Anomaly, on the other hand, is based on the extensive BDD100K dataset~\cite{Yu2018BDD100KAD}, where all instances from several classes were removed from the training and validation sets and thus treated as anomalies in a novel test set.
\subsubsection{Tasks and Ground truth}
\begin{figure}[t]
    \captionsetup[subfloat]{labelformat=empty}
    \centering
    \subfloat[StreetHazards]{\includegraphics[width=0.485\columnwidth]{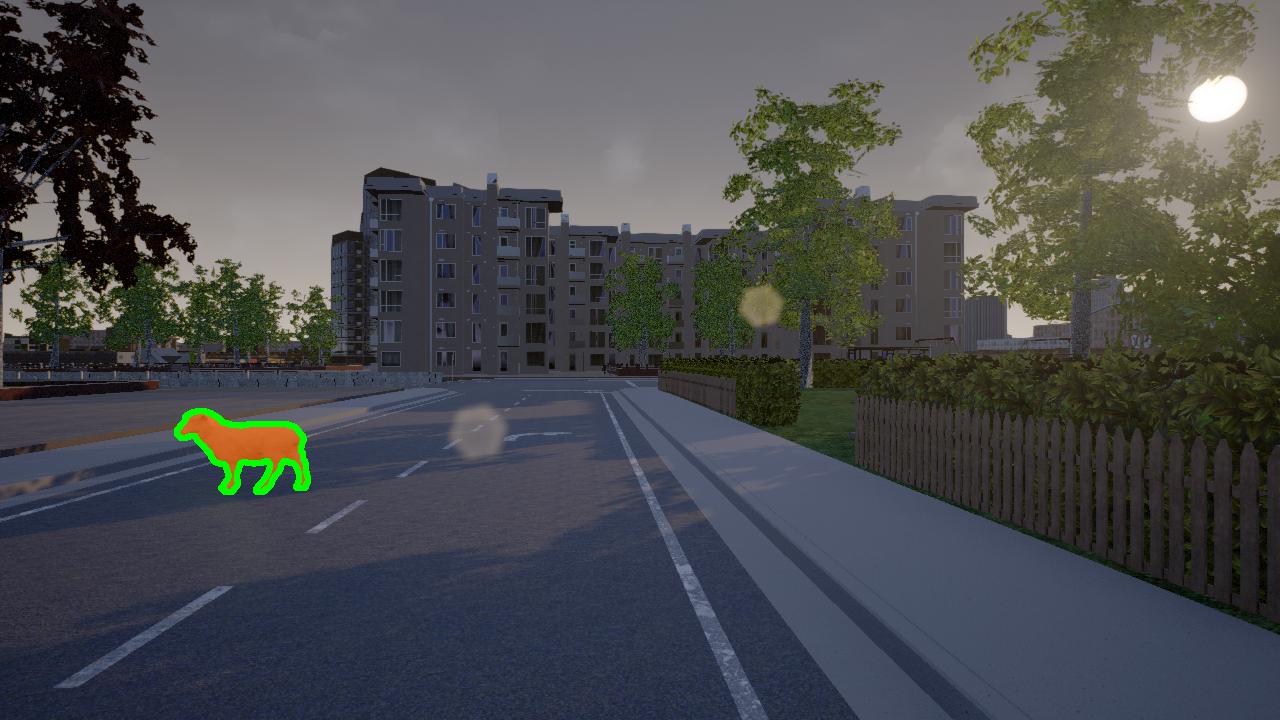}}\hfill
    \subfloat[BDD-Anomaly]{\includegraphics[width=0.485\columnwidth]{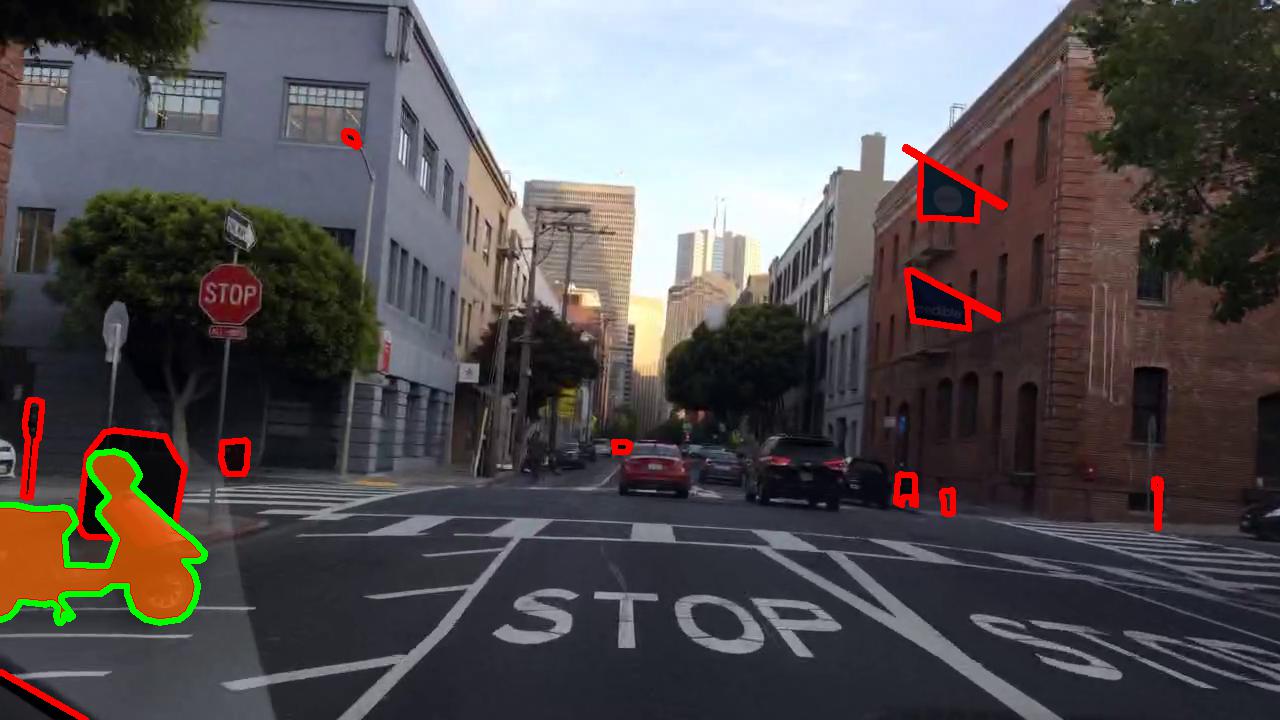}}
    \caption{CAOS: Simulated (left) and hypothetical (right) anomalies in the StreetHazards and BDD-Anomaly test sets.}
    \label{fig:caos}
\end{figure}

Both datasets are designed for the task of semantic segmentation and were aligned to provide both RGB image data and semantic ground truth in the same resolution. The StreetHazards dataset provides a wide variety of scenarios. In total, 250 different anomalies were inserted, taken from the Digimation Model Bank Library and semantic ShapeNet. A complete list can be found in~\cite{Hendrycks_Scaling_2022_ICML}. The semantic masks are fully annotated with one additional \textit{anomaly} class.
For BDD-Anomaly, the three classes \textit{motorcycle}, \textit{train}, and \textit{bicycle} are treated as anomalies. All frames from the original training and validation sets, which include these classes, were moved to the novel test set, where they became anomalies. An example is provided in Figure~\ref{fig:caos}. As the BDD100K dataset provides fully annotated semantic masks, they are also available for BDD-Anomaly, which makes the anomalous classes distinguishable. As the validation sets for both StreetHazards and BDD-Anomaly do not include anomalies, but are only provided for the regular task of semantic segmentation, we excluded them from Table~\ref{tab:overview}.

\subsubsection{Context}
For StreetHazards, the CARLA towns show slight variations but follow the same theme. Different weather and daytime settings are included. While the dataset provides scenarios, the ego-motion seems to have been performed manually, as it is rather inconsistent. Most anomalies are not placed in regions with relevance to the driving task. As BDD-Anomaly is based on BDD100K, it includes varying sceneries, weather conditions, and times of the day.

\subsubsection{License}
CAOS is provided under the MIT license~\cite{caos_web}. For BDD100K, “permission to use, copy, modify, and distribute this software and its documentation for educational, research, and not-for-profit purposes”~\cite{BDD_Yu_2022_Web} is granted.

\subsection{WD-Pascal} % 11/2019
WD-Pascal~\cite{Bevandic_Simultaneous_2019_GCPR} is a small dataset published in 2019, where the WildDash (WD) dataset~\cite{Zendel_2018_ECCV} was augmented with animals from the PASCAL VOC 2007 dataset~\cite{Everingham2010ThePV}.

\subsubsection{Tasks and Ground truth}
The dataset is produced for the task of semantic segmentation, but the data is not provided explicitly. As part of the author's code, it is assembled on the fly and provided as a PyTorch dataset~\cite{wdpascal_web}.

\subsubsection{Context} 
\begin{figure}[t]
    \captionsetup[subfloat]{labelformat=empty}
    \centering
    \subfloat{\includegraphics[trim={0 8px 0 0},clip,width=0.485\columnwidth]{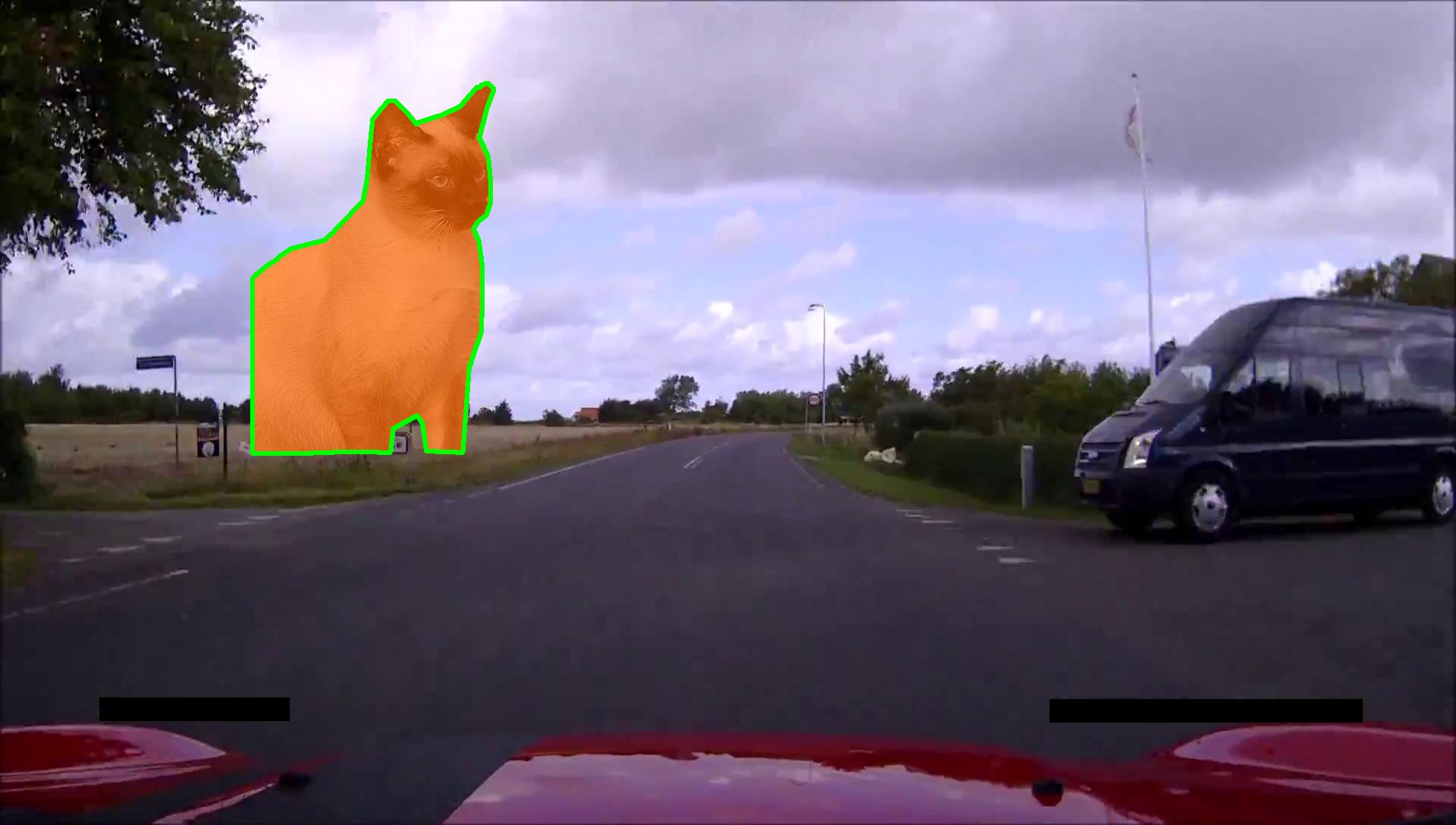}}\hfill
    \subfloat{\includegraphics[trim={0 8px 0 0},clip,width=0.485\columnwidth]{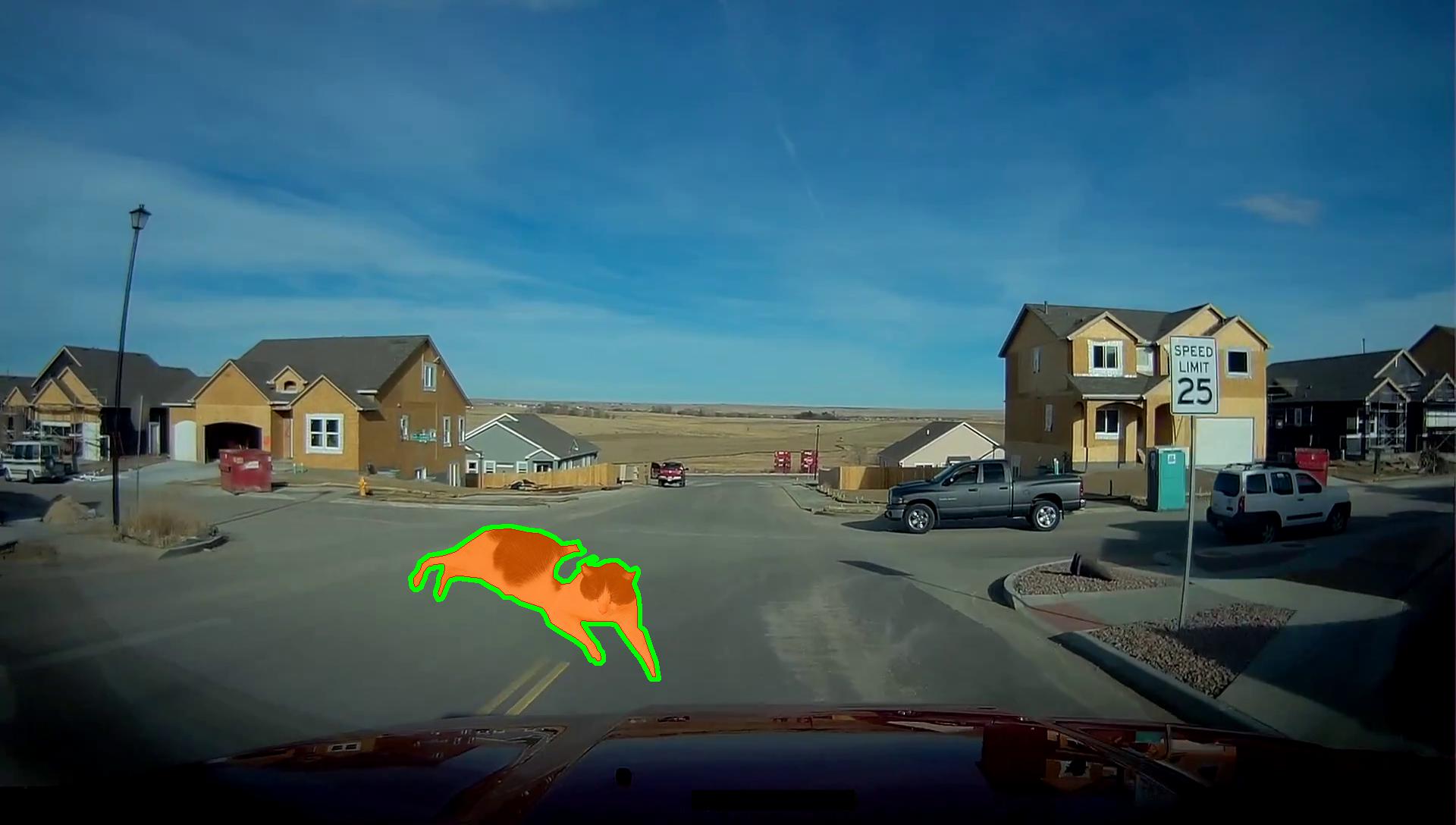}}
    \caption{WD-Pascal: Two examples of synthetically inserted anomalies into real-world scenes.}
    \label{fig:wdpascal}
\end{figure}

While the dataset is small, the variety remains relatively high due to the WildDash dataset, as shown in Figure~\ref{fig:wdpascal}. The included animals are not always complete and vary in size, which often leads to unrealistic augmentations.

\subsubsection{License}
The WD-Pascal generation code is provided under a GPL-2.0 license~\cite{wdpascal_web}. The necessary WildDash dataset comes with an extensive license agreement, where only the intensity images are released under the CC BY-NC 4.0 license~\cite{wilddash_web}. For the PASCAL VOC dataset, no license agreement is mentioned~\cite{voc_web}. However, some images are provided by Flickr, which introduce their own Terms of Use.

\subsection{Vistas-NP} % 2020
Vistas-NP~\cite{visapp21}, introduced in 2020, is a large-scale anomaly dataset based on the Mapillary Vistas dataset~\cite{neuhold_mapillary_2017}. Similarly to BDD-Anomaly, they have excluded classes from the train and validation splits, creating a novel test split with hypothetical anomalies. With over 11,000 labeled frames, Vistas-NP is the largest anomaly dataset to date.

\subsubsection{Tasks and Ground truth}
\begin{figure}[t]
    \captionsetup[subfloat]{labelformat=empty}
    \centering
    \subfloat{\includegraphics[trim={0 747px 0 0},clip,width=0.485\columnwidth]{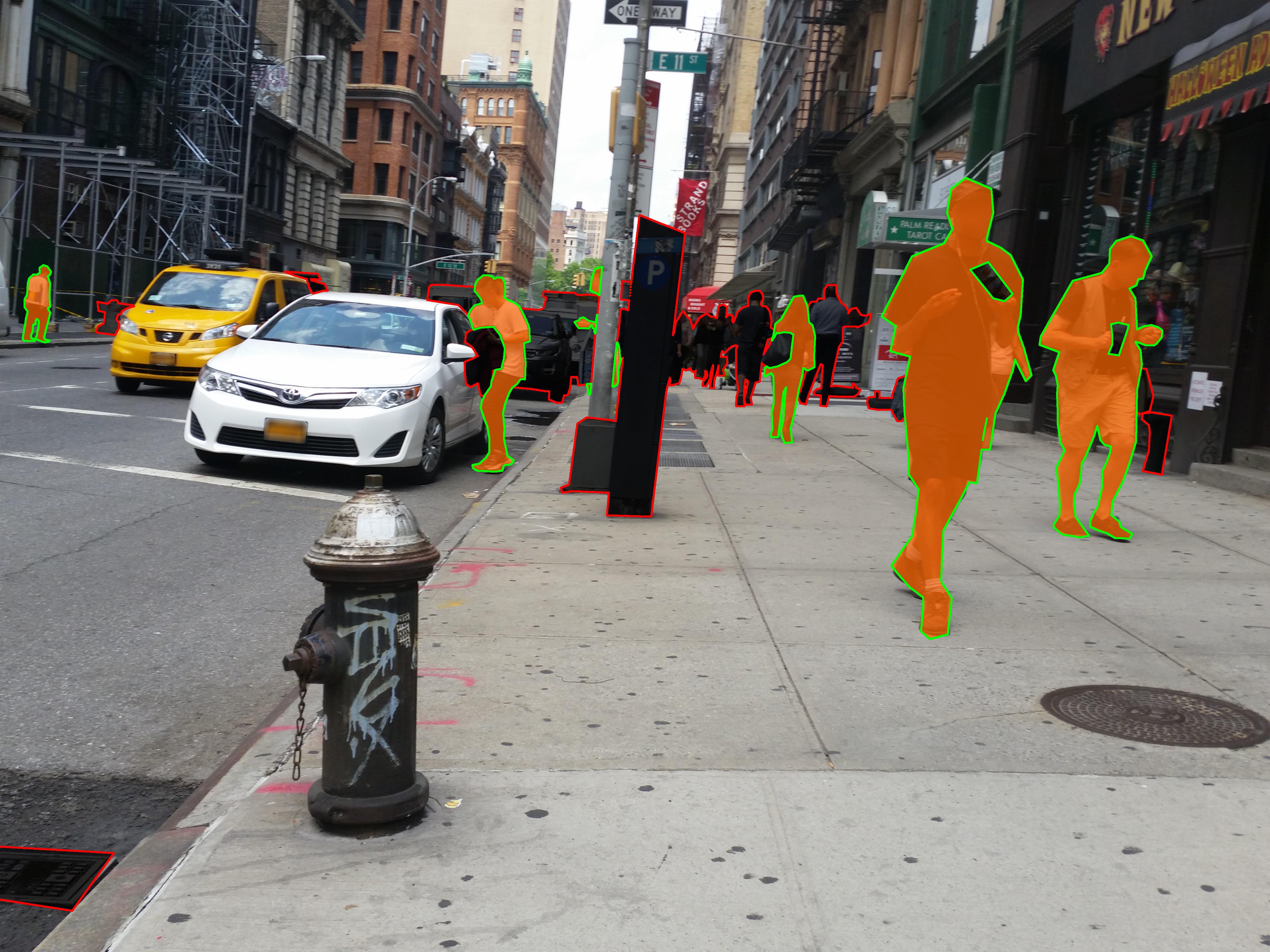}}\hfill
    \subfloat{\includegraphics[trim={0 312px 0 300px},clip,width=0.485\columnwidth]{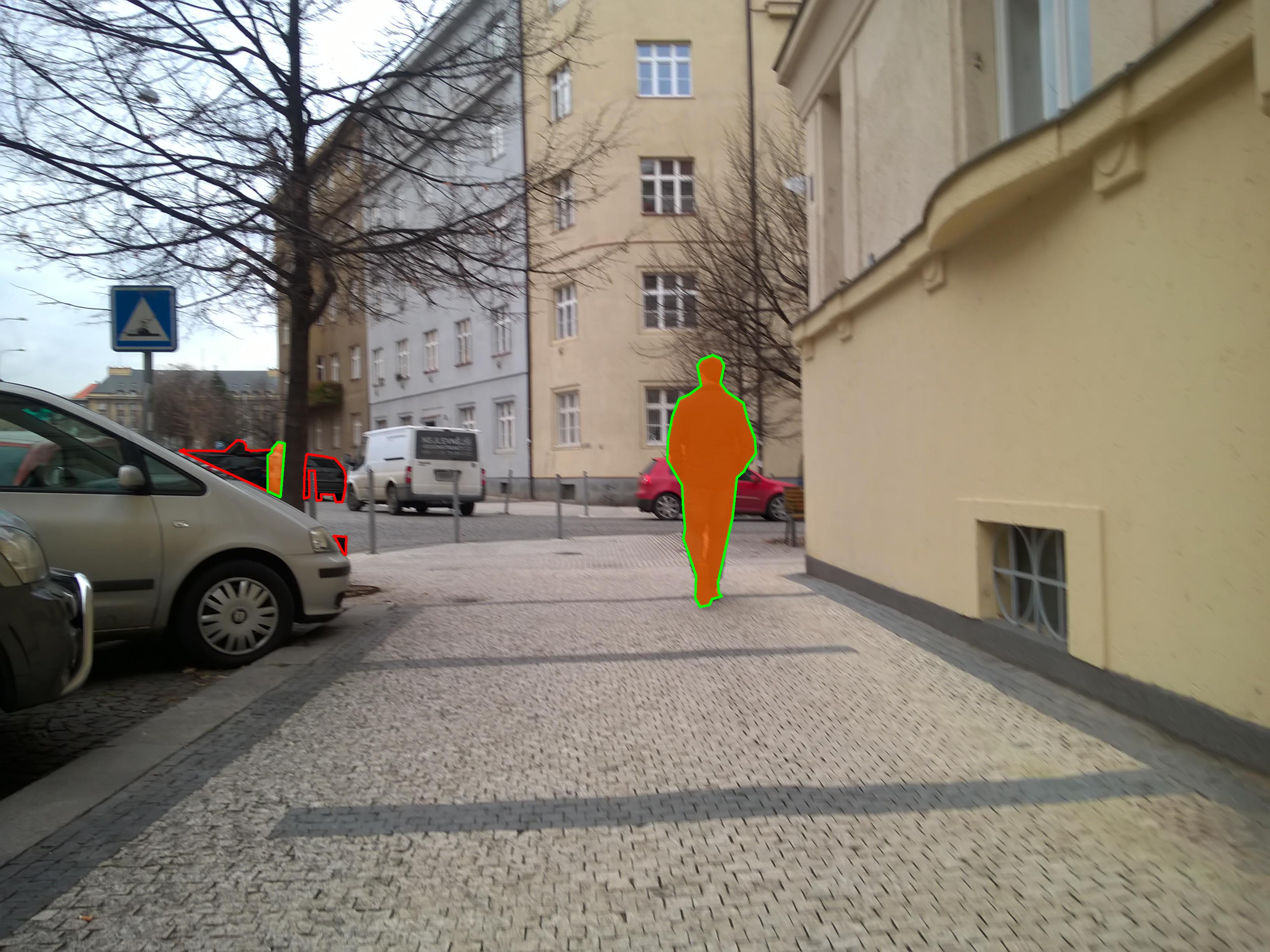}}
    \caption{Vistas-NP: Examples showing humans as hypothetical anomalies in the Vistas-NP test split.}
    \label{fig:vistasnp}
\end{figure}

The dataset is designed for the task of semantic segmentation, as visible in Figure~\ref{fig:vistasnp}. The chosen anomaly classes differ from those in BDD-Anomaly to avoid visual similarity of anomalous and non-anomalous classes, e.g., \emph{train} and \emph{bus}. Hence, a whole category is excluded, which includes all classes associated with humans.

\subsubsection{Context}
The underlying Mapillary Vistas dataset has a large variety. Compared to BDD-Anomaly, where all images originate from the USA, images from multiple countries are included. As a crowdsourcing approach is utilized, this is reflected in a wide variety of resolutions.

\subsubsection{License}
The “Vistas-NP dataset should be used under the same conditions as the original dataset”~\cite{vistasnp_web}, which is provided under the CC BY-NC-SA 4.0 license~\cite{mapillary_web}.

\subsection{SegmentMeIfYouCan} % 2021
The SegmentMeIfYouCan benchmark~\cite{Chan_SegmentMeIfYouCan_2021_NEURIPS} was developed in 2021 by Chan et al., introducing two real-world datasets. A previous version of the RoadAnomaly21 dataset was already published in 2019 by Lis et al.~\cite{Lis2019DetectingTU}. The current version was both refined and extended. It consists of images collected from the internet, which show anomalous objects on or near the road. The RoadObstacle21 dataset was recorded by the authors and includes anomalous objects placed on the road ahead. Similar to FS Lost and Found, which is also included in the benchmark, these datasets only contain real anomalies.

\subsubsection{Tasks and Ground truth}
\begin{figure}[t]
    \captionsetup[subfloat]{labelformat=empty}
    \centering
    \subfloat[RoadAnomaly21]{\includegraphics[width=0.485\columnwidth]{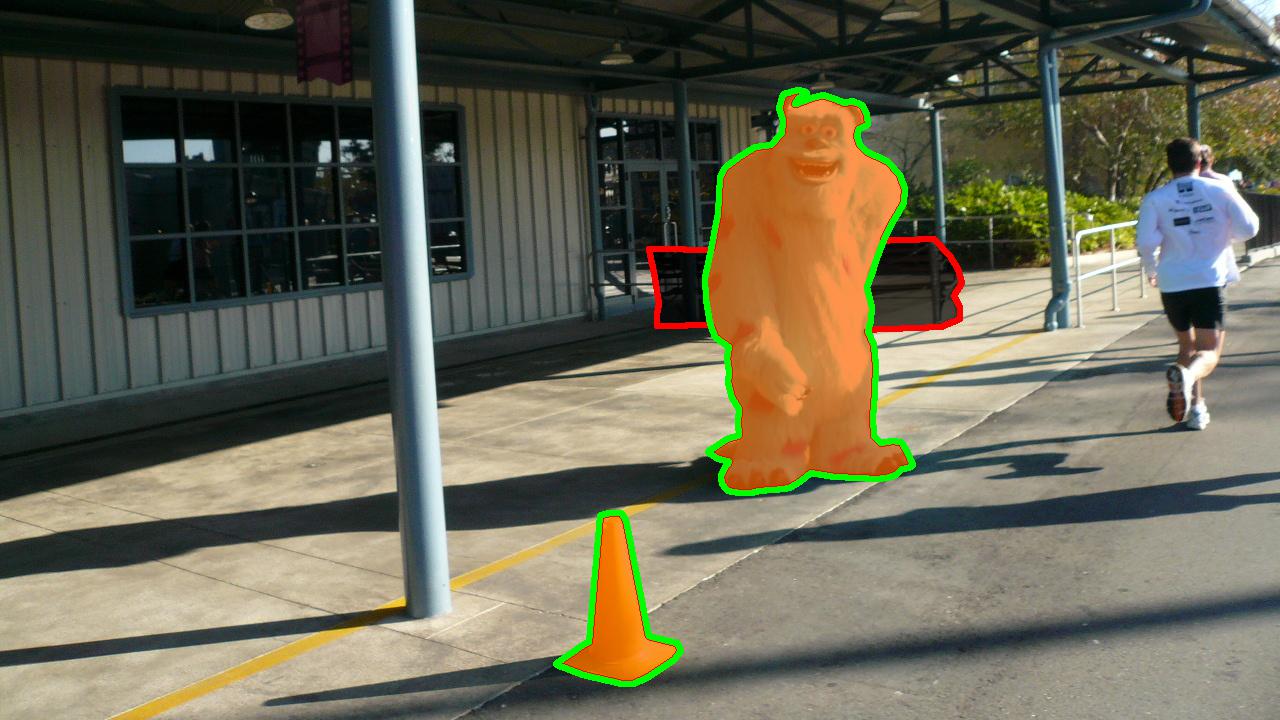}}\hfill
    \subfloat[RoadObstacle21]{\includegraphics[width=0.485\columnwidth]{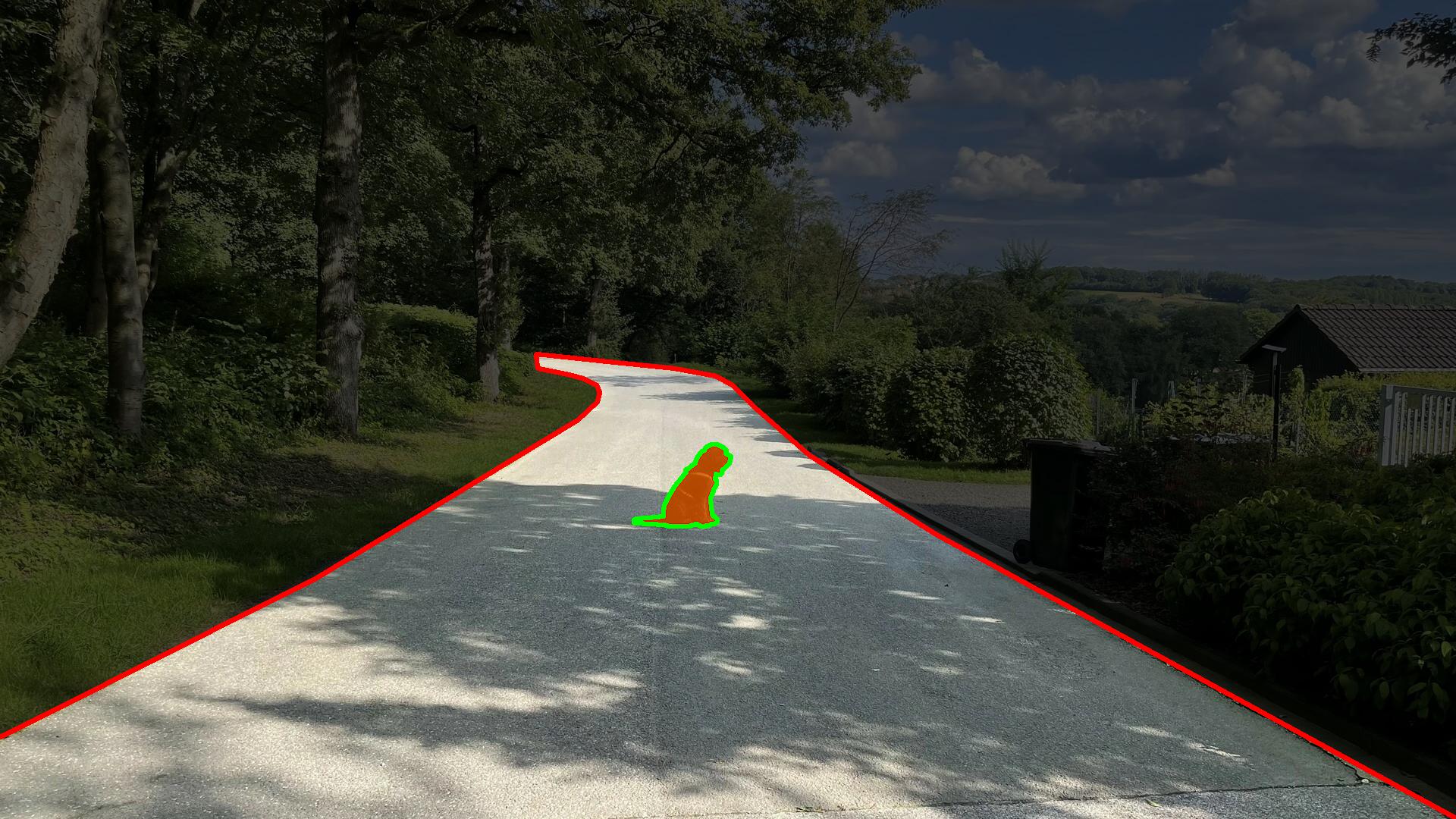}}
    \caption{SegmentMeIfYouCan: Real-world examples from the RoadAnomaly21 test and the RoadObstacle21 val splits.}
    \label{fig:smiyc}
\end{figure}

Both datasets are designed for the task of semantic segmentation, the semantic masks include binary anomaly labels. RoadAnomaly21 is designed for general anomaly detection in full street scenes, whereas in RoadObstacle21, the road is considered the region of interest, i.e., the \emph{not anomaly} class. Thus, everything not included in this region is assigned to the \emph{void} class, which is represented in Figure~\ref{fig:smiyc}. The anomalies in RoadAnomaly21 can be categorized into animals, e.g., \emph{elephant, cow, horse}, unknown vehicles, e.g., \emph{airplane, boat trailer, tractor}, and others, such as \emph{tent, piano}, or \emph{cones}. In RoadObstacle21, each object on the road ahead is considered an obstacle. However, all obstacles in this dataset also fit the definition of anomaly as objects which cannot be assigned to the Cityscapes classes. Semantic masks are in both cases only published for small validation sets.

\subsubsection{Context}
In RoadAnomaly21, images are collected from web resources and thus depict a wide variety of environments and settings. All images are recorded during daytime and in clear weather. The anomalies can appear anywhere in the image, even in the sky. Therefore, they are not necessarily street hazards. The images of RoadObstacle21 are recorded in Germany and Switzerland on seven different road types, also during daytime and in clear weather. Additionally, there are $55$ annotated frames by night and in snowy weather conditions. 
% All frames were extracted from video sequences, which were recorded while driving towards the obstacles. 
% Accordingly, the dataset contains several frames per sequence, showing the obstacles from different distances. 

\subsubsection{License}
RoadObstacle21 is provided under the CC BY 4.0, RoadAnomaly21 under different CC BY licenses.

\subsection{CODA} % 03/2022

The CODA datasets, released in 2022, are the first ones that are based primarily on datasets that include not only camera but also lidar data. They are divided into the CODA Base and the CODA2022 subsets, which in turn consist of different underlying datasets, namely KITTI~\cite{Geiger2012CVPR}, nuScenes~\cite{nuscenes}, ONCE~\cite{mao2021one}, and SODA10M~\cite{han2021soda10m}. CODA Base was described in~\cite{Li_CODA_2022_ECCV}, while CODA2022 was a later addition~\cite{coda_web}.

\subsubsection{Tasks and Ground truth}
\begin{figure}[t]
    \captionsetup[subfloat]{labelformat=empty}
    \centering
    \subfloat[CODA-nuScenes]{\includegraphics[width=0.485\columnwidth]{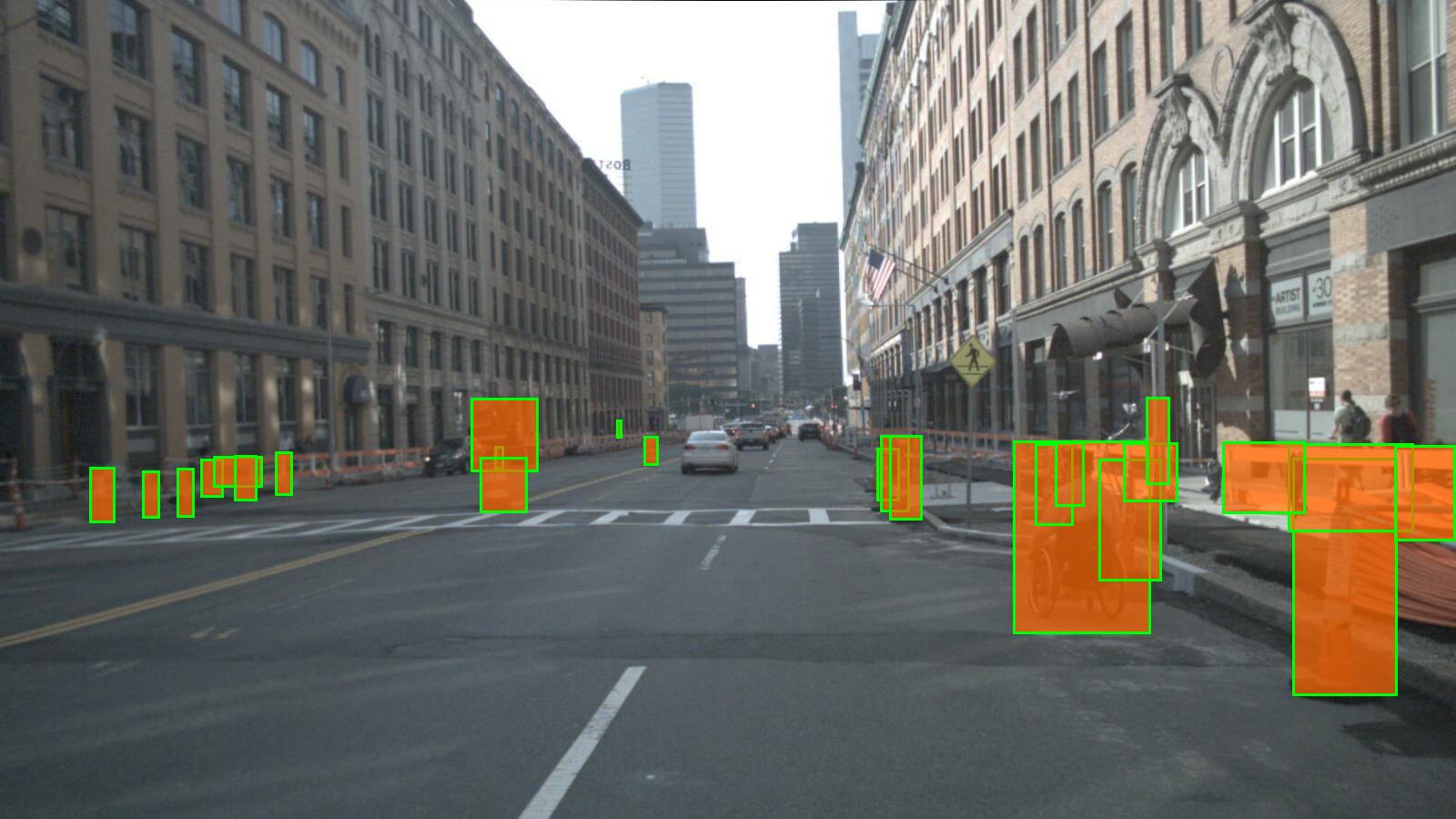}}
    \hfill
    \subfloat[CODA-ONCE]{\includegraphics[trim={0 0 107px 0},clip,width=0.485\columnwidth]{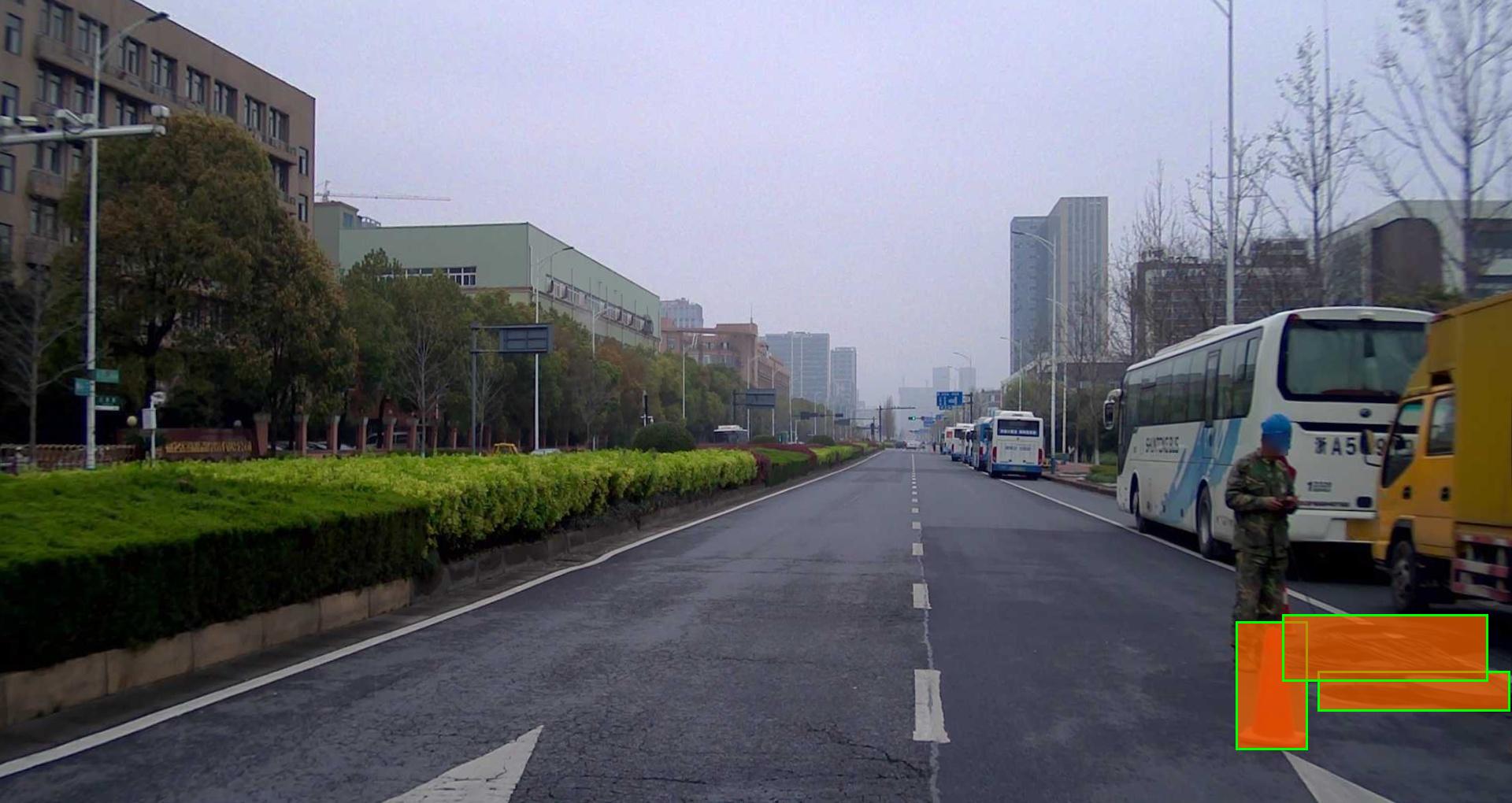}}\hfill
    \subfloat[CODA-KITTI]{\includegraphics[width=\columnwidth]{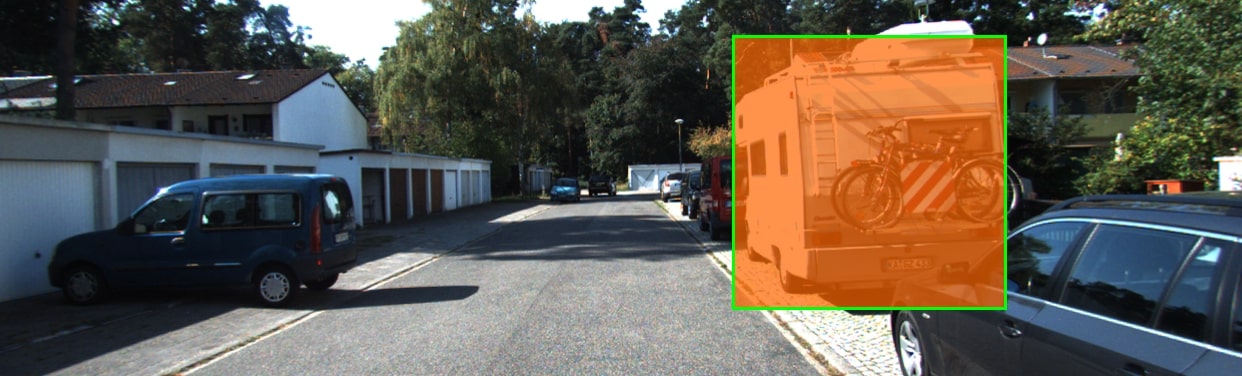}}\hfill
    \subfloat[CODA2022-SODA10M]{\includegraphics[width=0.485\columnwidth]{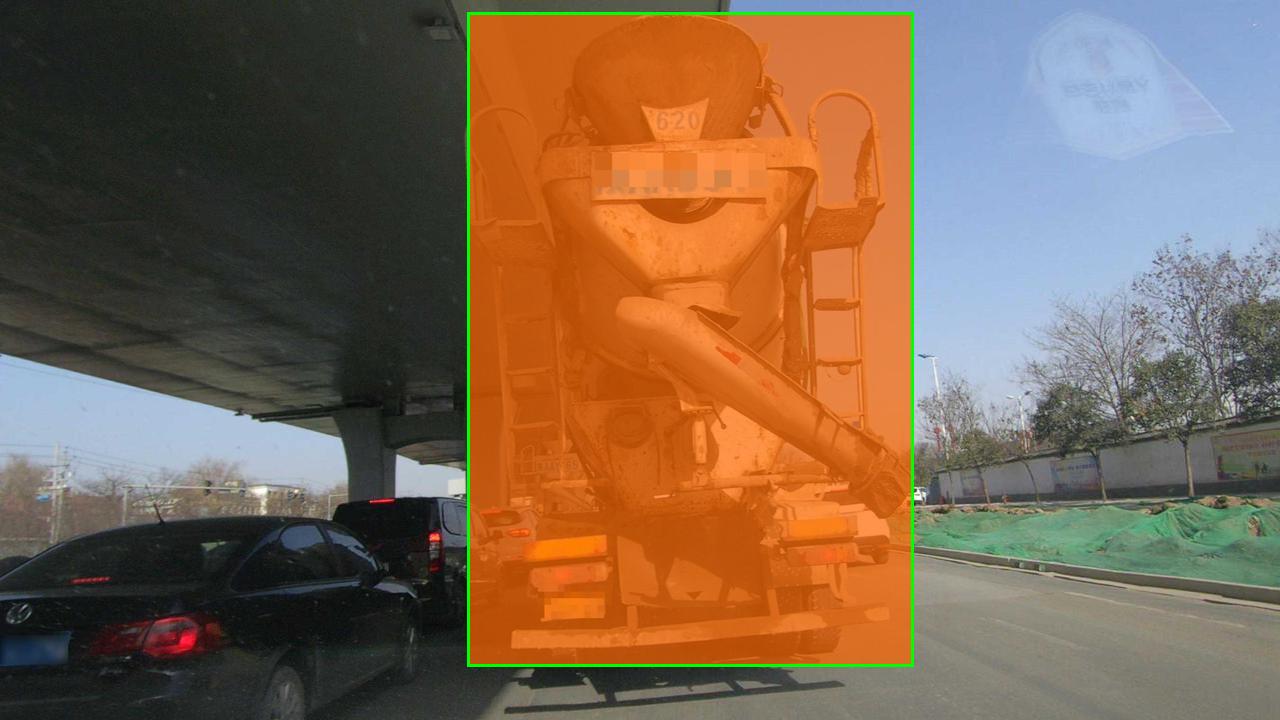}}\hfill
    \subfloat[CODA2022-ONCE]{\includegraphics[trim={0 0 75px 0},clip,width=0.485\columnwidth]{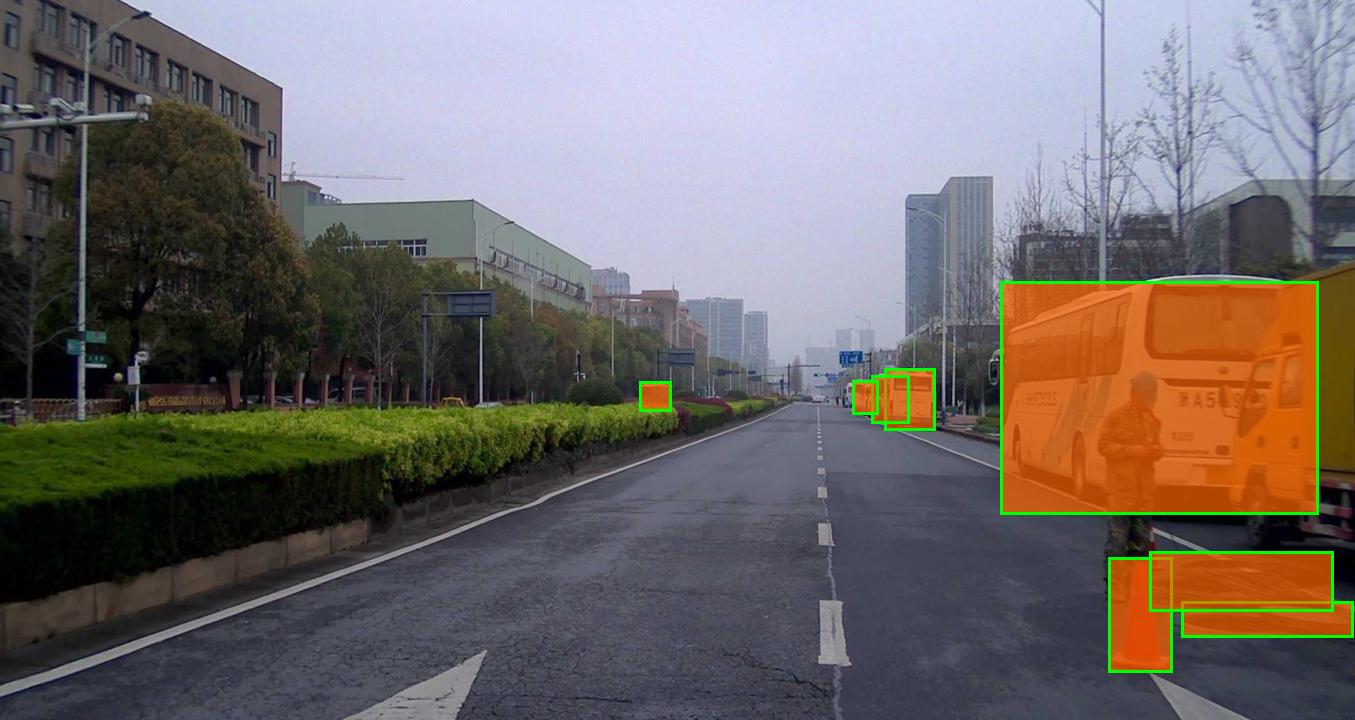}}
    \caption{CODA: Annotation of real-world anomalies from CODA Base and CODA2022, based on different techniques.}
    \label{fig:codabase}
\end{figure}

The CODA datasets are designed for object detection with ground truth anomaly bounding boxes only in the image space, as Figure~\ref{fig:codabase} shows. Common objects are only labeled in the CODA2022 datasets.
For the labeling of anomalies, different techniques were used. For CODA-ONCE, unknown clusters from the lidar space were mapped to the image space and those, that could not be classified by an object detector, remained as anomaly proposals. In a second stage, a manual process was applied to label the images. This includes proposal refinement, false positive removal and manual additions. Additionally, pre-labeling with CLIP~\cite{pmlr-v139-radford21a} was used. Two types of anomalies were considered: Risky objects, which might block the ego-vehicle, or novel objects, which do not belong to a typical category. For CODA-KITTI and CODA-nuScenes, only the second stage was applied, where uncommon classes from the existing labels were used as proposals, e.g., the \textit{misc} category in KITTI. The anomalies are grouped into the categories \textit{vehicle}, \textit{pedestrian}, \textit{cyclist}, \textit{animal}, \textit{traffic facility}, \textit{obstruction}, and \textit{misc}. For CODA2022, a pre-labeling process based on FILIP~\cite{filip_2021} was employed, followed by a manual process, as SODA10M is an unlabeled dataset without lidar data. The dataset is in general larger and includes more anomaly categories. The comparisons in Figure~\ref{fig:codabase} and Figure~\ref{fig:anomalymasks} clearly show the differences for the CODA-ONCE and CODA2022-ONCE datasets.

\subsubsection{Context}
As the CODA datasets are based on four different datasets, their variety is rather high, including multiple countries, weather conditions, and times of day. For most of the datasets, the majority of the anomalies appear on the side of the road with little relation to the ego path. Only in CODA2022-SODA10M, a different picture emerges, where more anomalies can be found in the center of the images. This is due to the crowdsourcing approach of the dataset.
\subsubsection{License}
As ONCE, SODA10M and CODA are all related to Huawei, their images are included in the datasets, while KITTI and nuScenes need to be downloaded separately. The CODA dataset is provided under the CC BY-NC-SA 4.0 license~\cite{coda_web}. KITTI is available under the CC BY-NC-SA 3.0 license, and nuScenes under the CC BY-NC-SA 4.0 license.

\subsection{Wuppertal OOD Tracking} 
These datasets were introduced in 2022 by Maag et al., enabling OOD detection and tracking over video sequences. The Street Obstacle Sequences (SOS) dataset contains annotated real-world scenes with real anomalies. The CARLA-WildLife (CWL) dataset is a synthetic dataset similar to StreetHazards, where freely available assets were inserted as anomalies. A third dataset, Wuppertal Obstacle Sequences (WOS), consists of real-world, but unlabeled sequences.

\subsubsection{Tasks and Ground truth}
SOS and CWL provide labels for the tasks of semantic segmentation, instance segmentation and depth estimation. The datasets include semantic masks with binary as well as class-specific anomaly labels. Analogously to RoadObstacle21, the road represents the region of interest, thus, everything besides the road is assigned to the \emph{void} class. Furthermore, both datasets include instance and depth masks. For SOS, 1,129 out of the 8,994 total frames were manually labeled. For CWL, also pixel-wise distance masks and fully annotated semantic masks are available. The anomalies in SOS belong to $13$ anomaly types, e.g., \emph{bag, umbrella} or \emph{toy}, the anomalies in CWL to $18$ anomaly types, e.g., \emph{dogs, pylons} or \emph{bags}. 

\subsubsection{Context}
\begin{figure}[t]
    \captionsetup[subfloat]{labelformat=empty}
    \centering
    \subfloat[CARLA-WildLife]{\includegraphics[width=0.485\columnwidth]{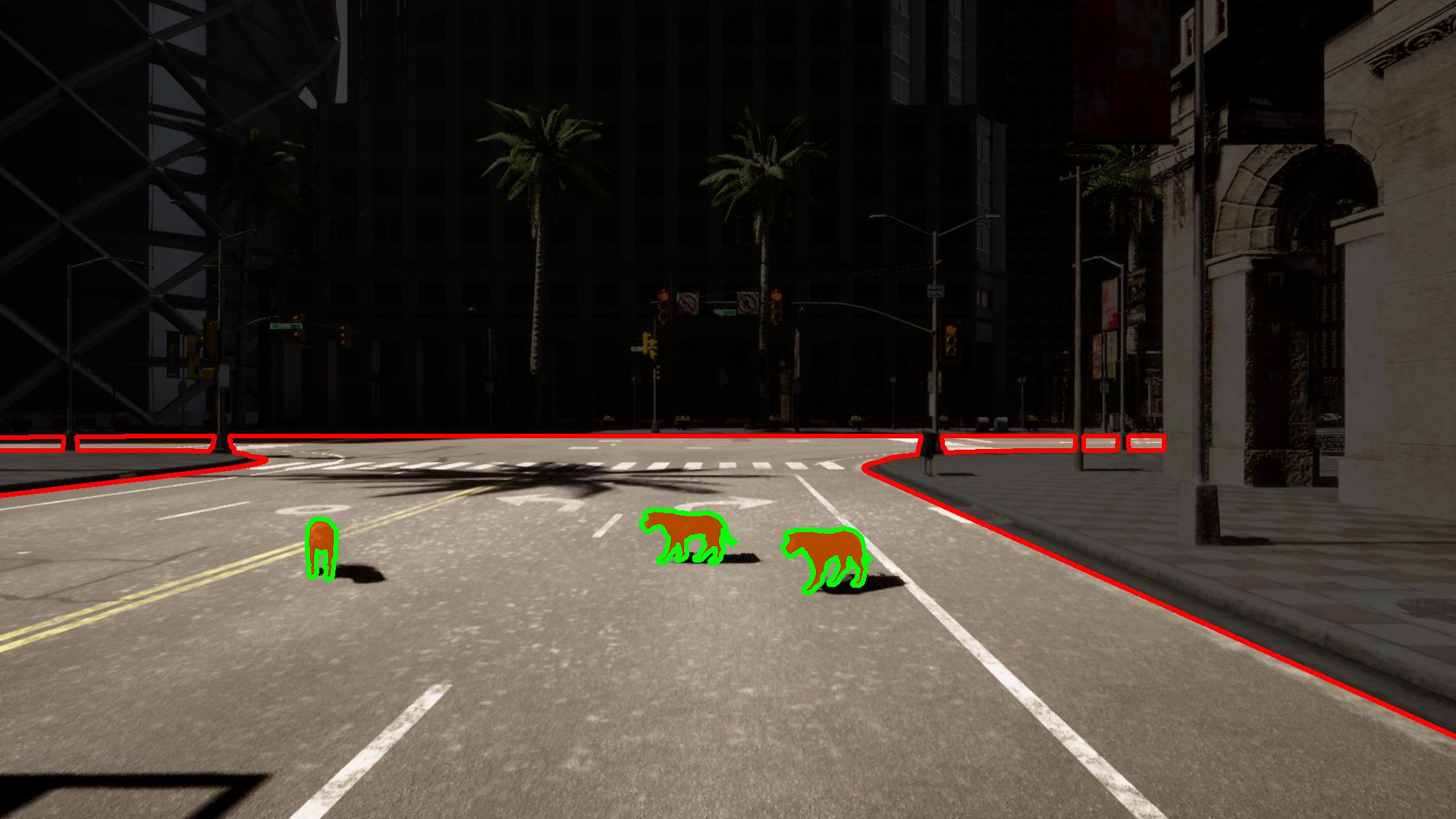}}\hfill
    \subfloat[Street Obstacle Sequences]{\includegraphics[width=0.485\columnwidth]{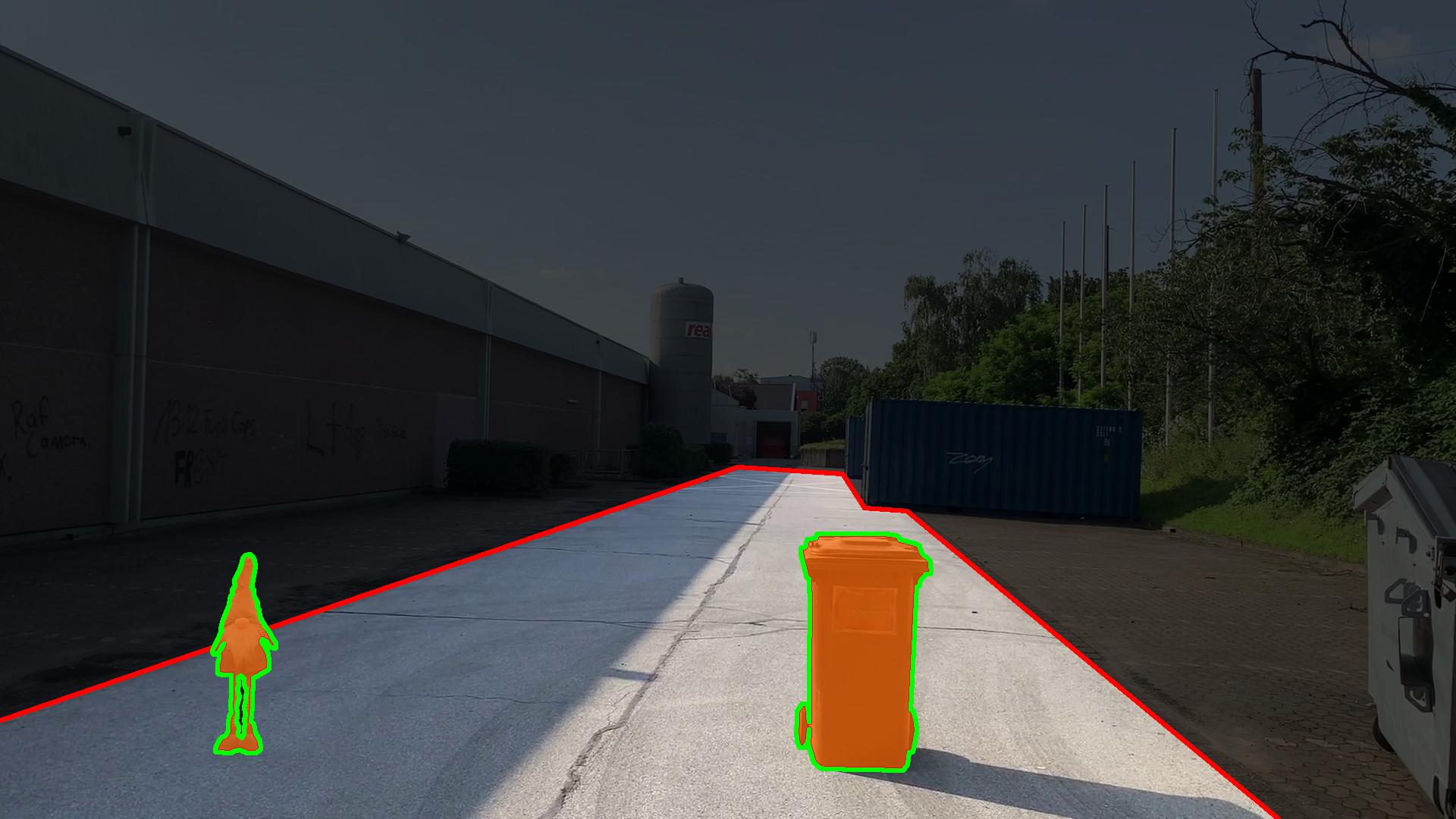}}
    \caption{Wuppertal OOD Tracking: Two examples showing simulated (left) and real (right) anomalies.}
    \label{fig:ood}
\end{figure}

All images, real and synthetic, were recorded during daytime in clear weather. The static objects in SOS and CWL are placed in a way that they are mostly enclosed by the road as shown in Figure~\ref{fig:ood}. In contrast to StreetHazards, the anomalies in CWL are all placed on the road ahead, which could cause safety-critical street scenarios. Furthermore, they are more realistic and of higher quality thanks to a newer version of the Unreal Engine. Besides the ego-vehicle and the anomalies, there are no other vehicles or humans in the scenes. 

\subsubsection{License}
All datasets are provided under CC BY 4.0 licenses. CARLA-WildLife was created using the Unreal Engine along with CARLA~\cite{Dosovitskiy2017CARLAAO}, provided under the MIT license. The assets from Unreal Engine 4.26, which were inserted as anomalies, are provided under CC BY licenses.

\section{Discussion}
\label{sec:conclusion}

We have presented a complete overview of 16 perception datasets in the field of anomaly detection for autonomous driving. We presented many techniques to define and generate anomalies, which come with certain challenges. In the following, we provide an extensive discussion of these.

\textbf{Definition of Normality.}
There is no clear definition of whether an object is anomalous or not. However, a common approach is to define anomalies as \emph{none-of-the-known classes} with respect to the $19$ Cityscapes evaluation classes. Most anomaly techniques adhere to the definition of “Cityscapes as normality”: For web sourcing, simulation, data augmentation and recording, the anomalous objects are selected to fit into this definition. For the void class approaches, the definition of anomalous objects depends on the respective underlying dataset, as the \emph{void} or \emph{misc} category isn't clearly defined, either. Also for the automated OOD proposal technique, the definition of normality strongly depends on the underlying classes of the employed detector(s). Finally, for class exclusion, normality depends on the choice of excluded classes. The anomalies labeled by this approach are usually not anomalous with respect to Cityscapes and as such do not represent anomalies which would be rare in the real world.

\begin{figure*}
    \captionsetup[subfloat]{labelformat=empty}
    \centering
    \subfloat[Lost and Found]{\includegraphics[height=2.0cm]{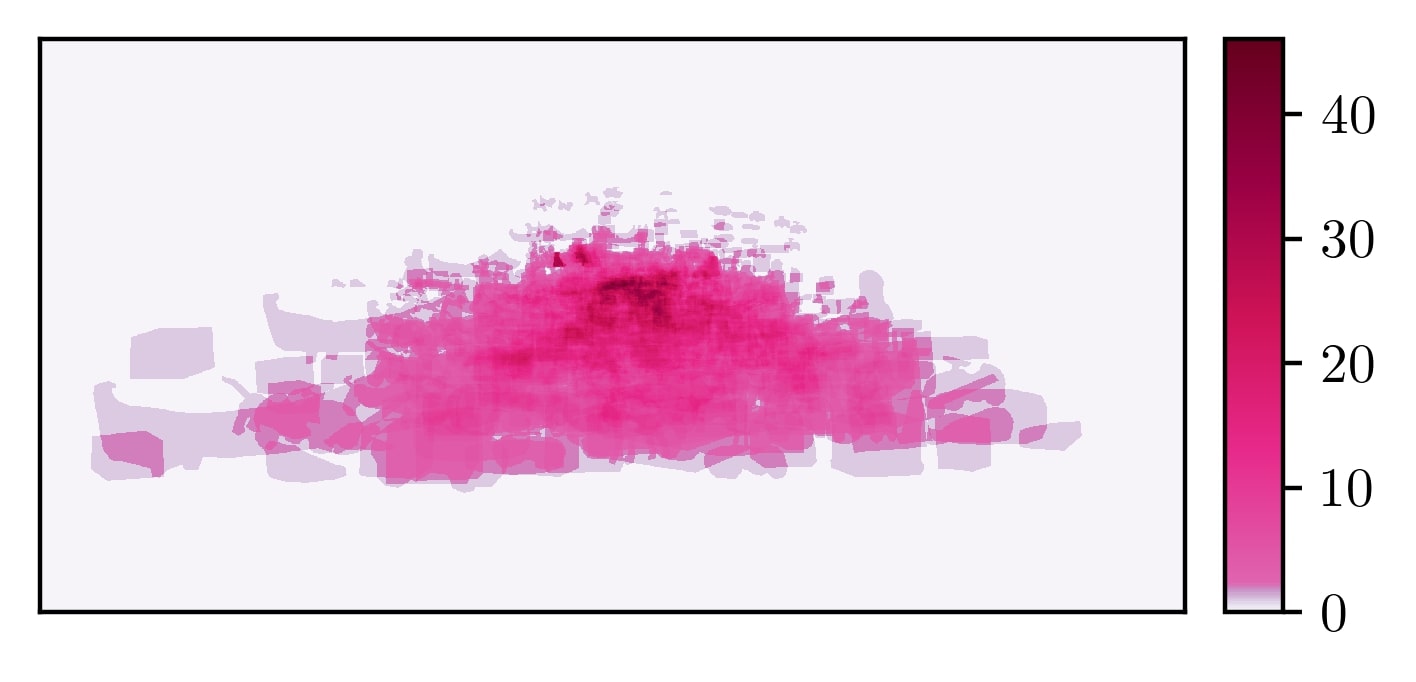}}\hfill
    \subfloat[FS Lost and Found (val)]{\includegraphics[height=2.0cm]{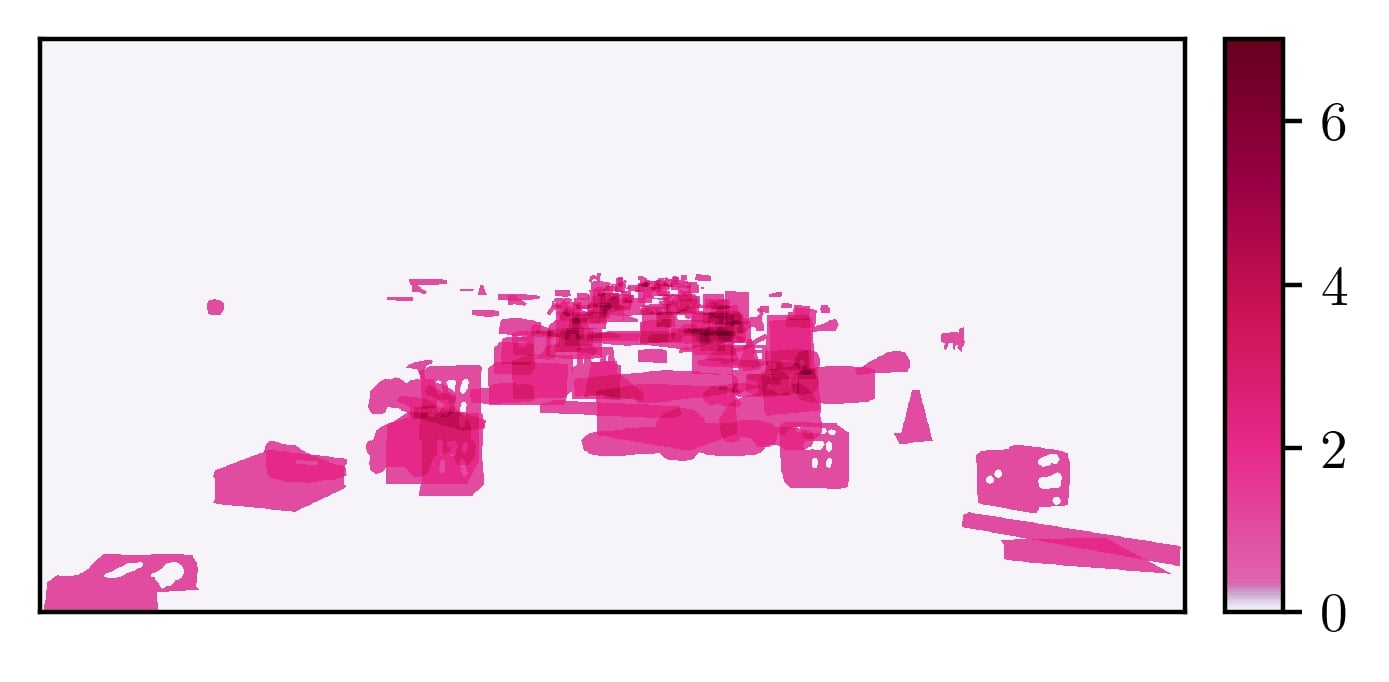}}\hfill
    \subfloat[FS Static (val)]{\includegraphics[height=2.05cm]{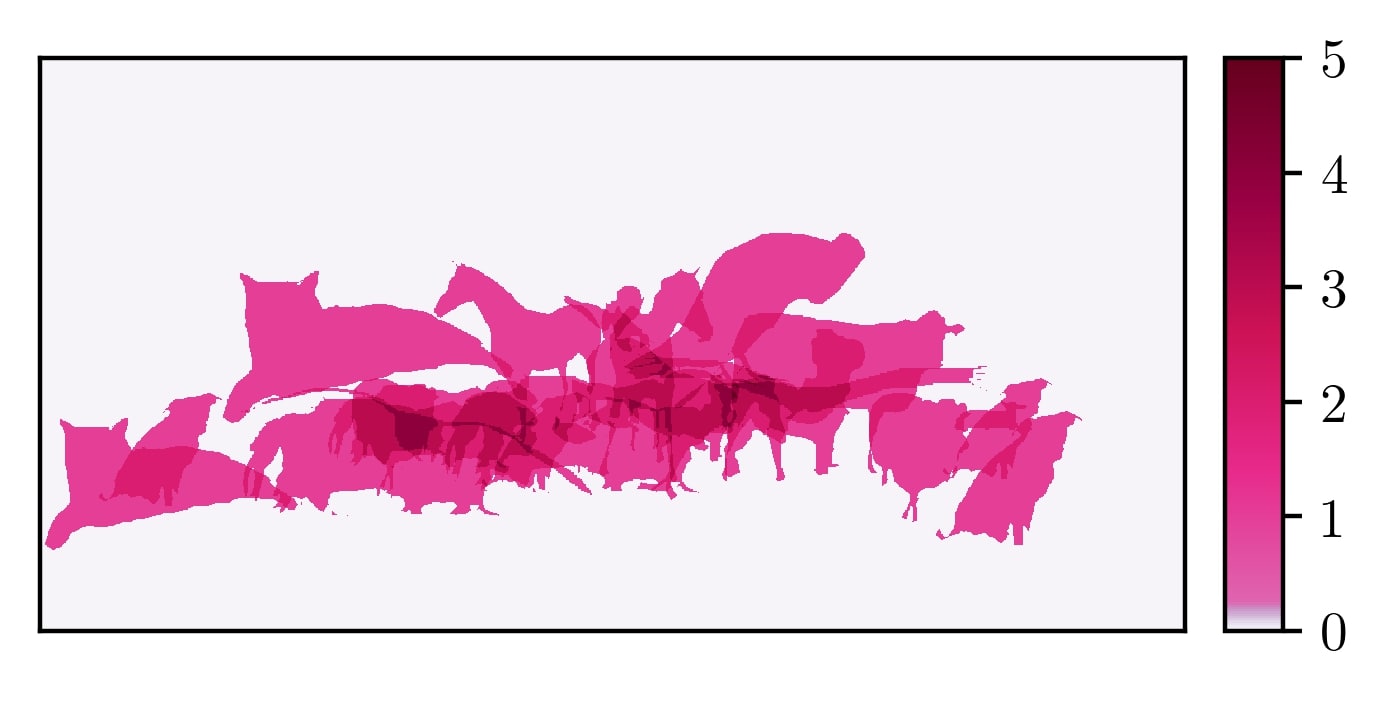}}\hfill
    \subfloat[StreetHazards]{\includegraphics[height=2.0cm]{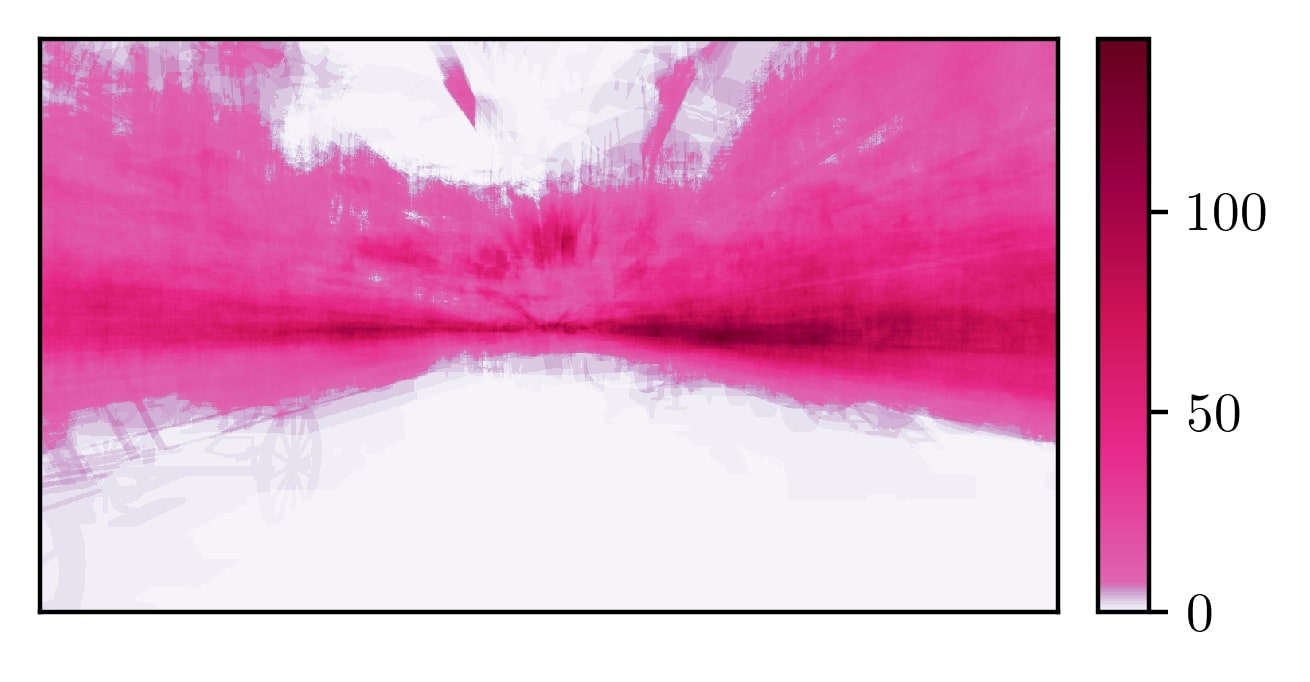}}\hfill
    \subfloat[BDD-Anomaly]{\includegraphics[height=2.0cm]{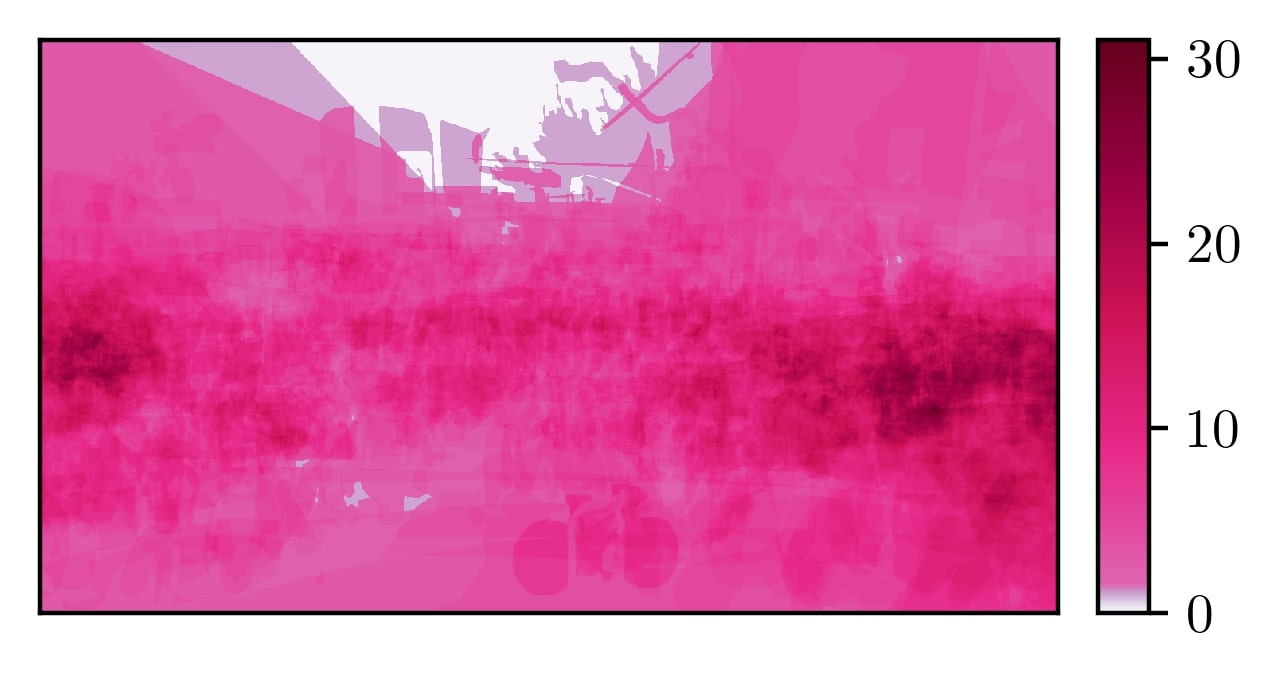}}\hfill
    \subfloat[WD-Pascal]{\includegraphics[height=2.0cm]{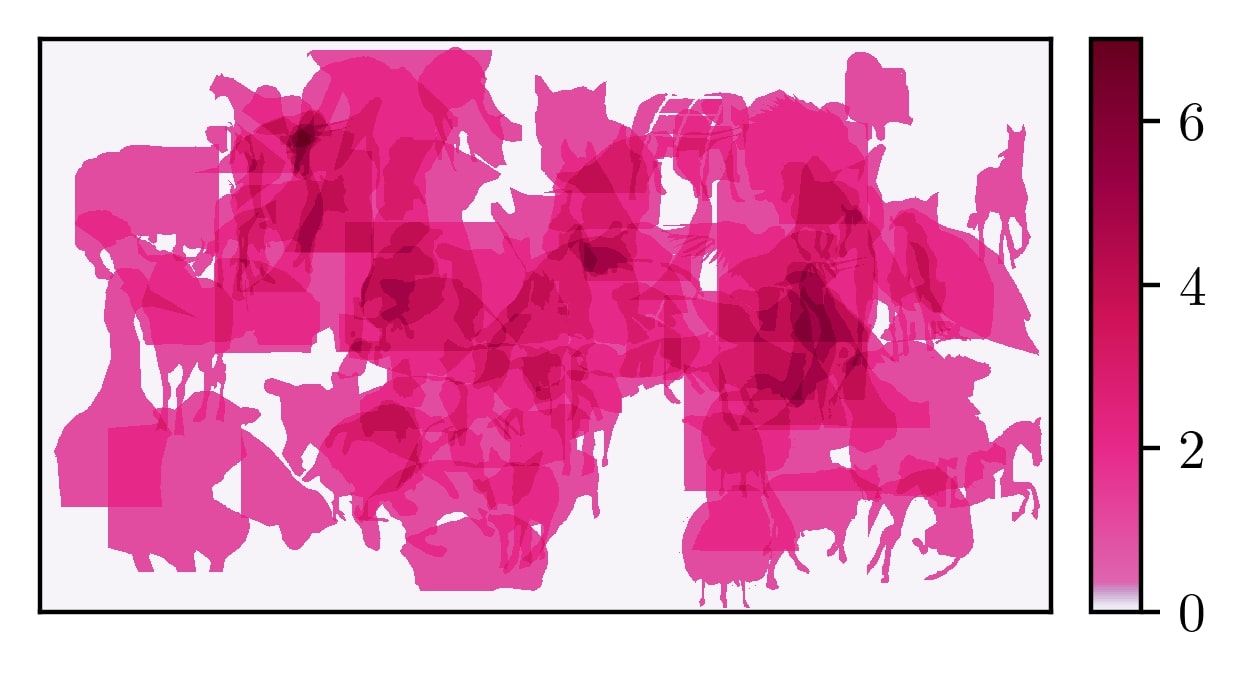}}\hfill
    \subfloat[Vistas-NP]{\includegraphics[height=2.0cm]{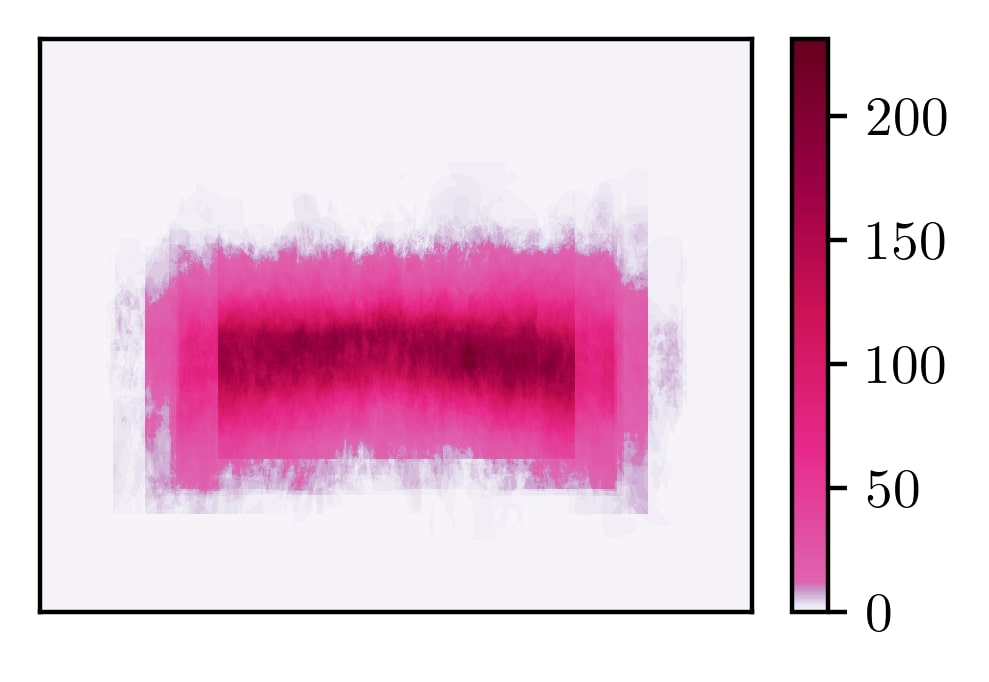}}\hfill
    \subfloat[RoadAnomaly21]{\includegraphics[height=2.0cm]{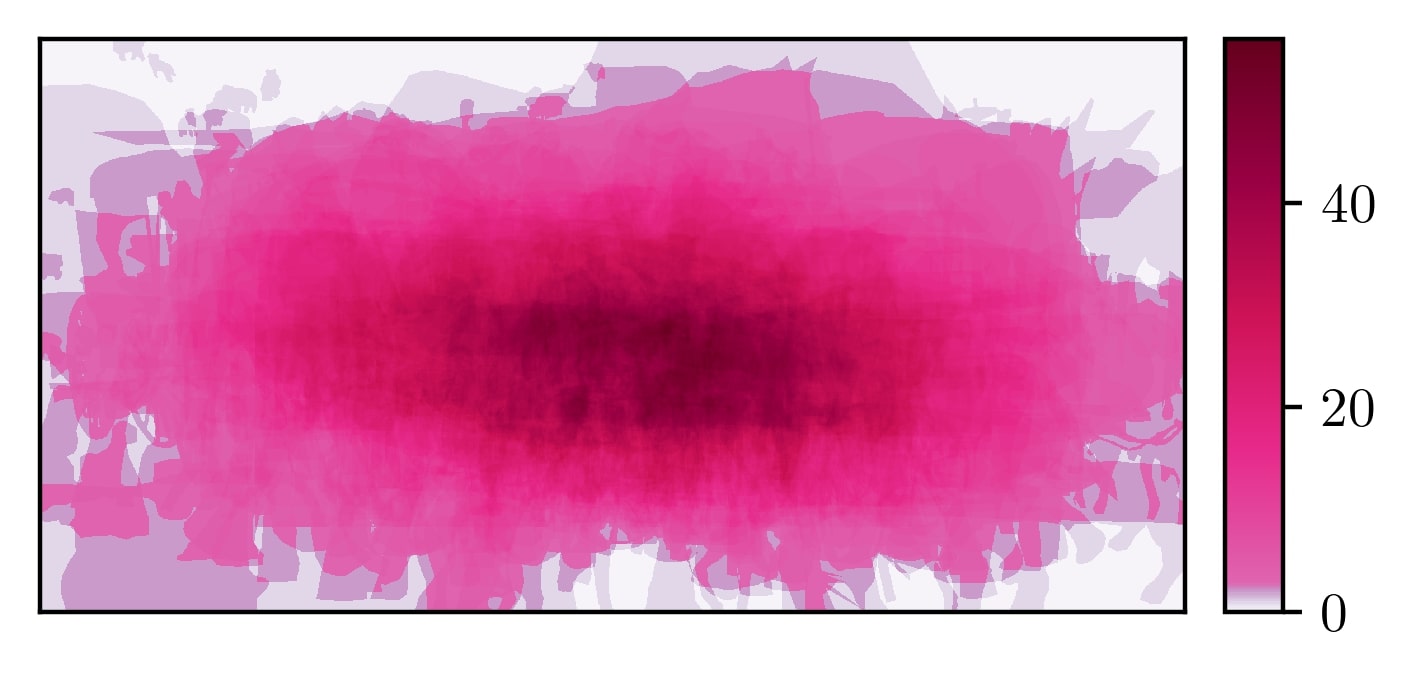}}\hfill
    \subfloat[RoadObstacle21]{\includegraphics[height=2.0cm]{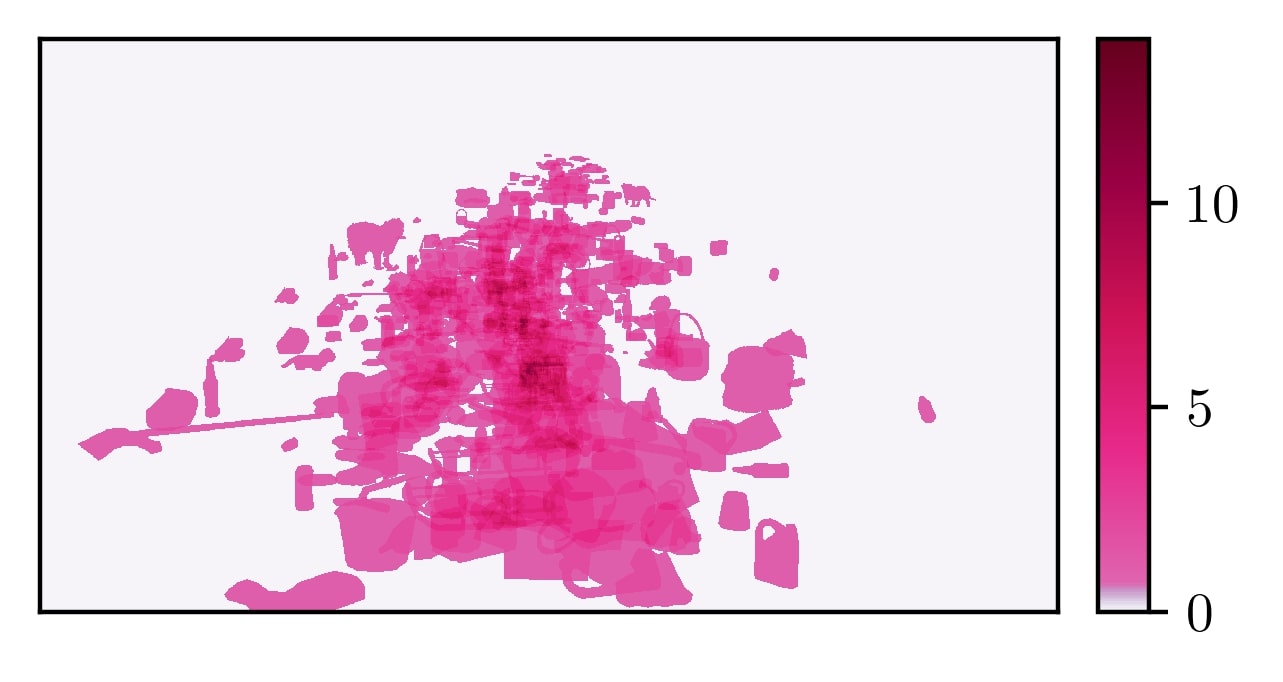}}\hfill
    \subfloat[CODA-KITTI]{\includegraphics[height=2.0cm]{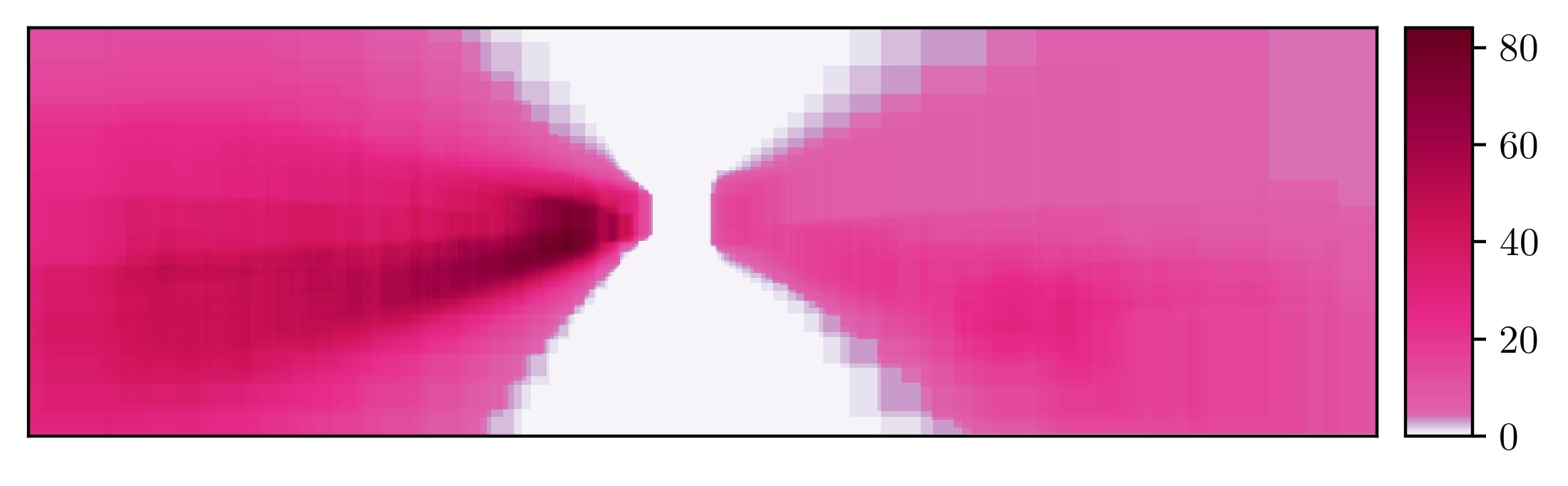}}\hfill
    \subfloat[CODA-nuScenes]{\includegraphics[height=2.03cm]{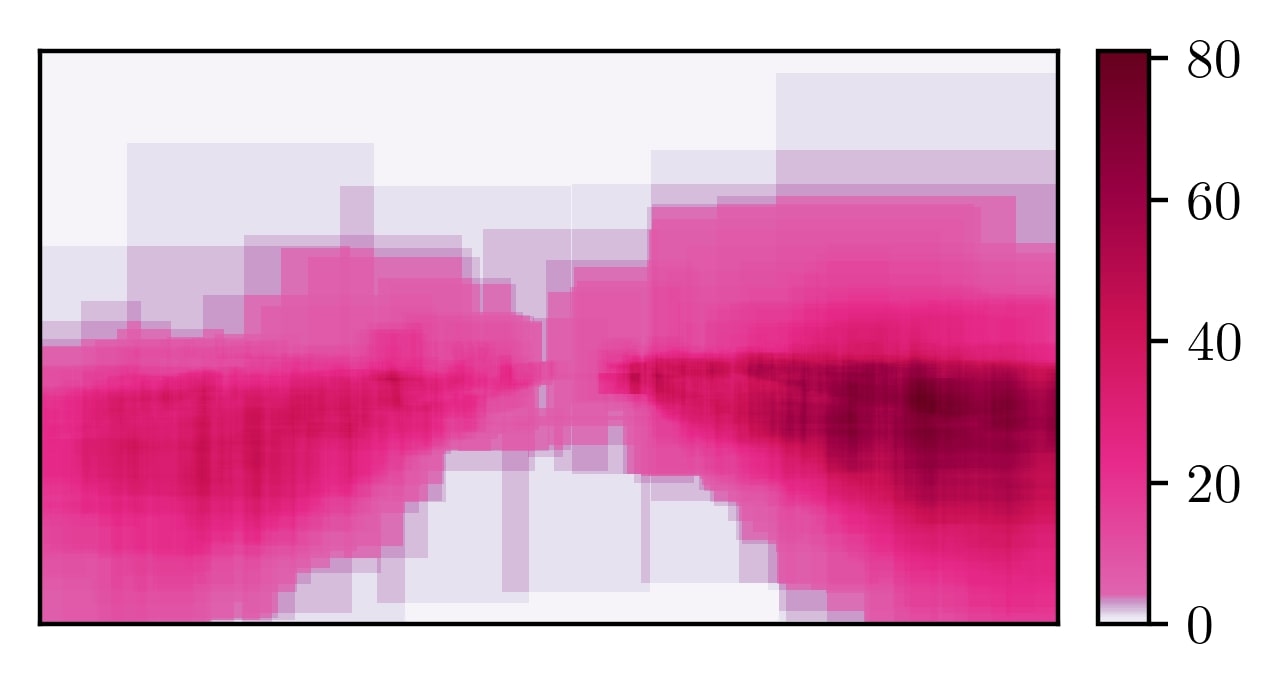}}\hfill
    \subfloat[CODA-ONCE]{\includegraphics[height=2.0cm]{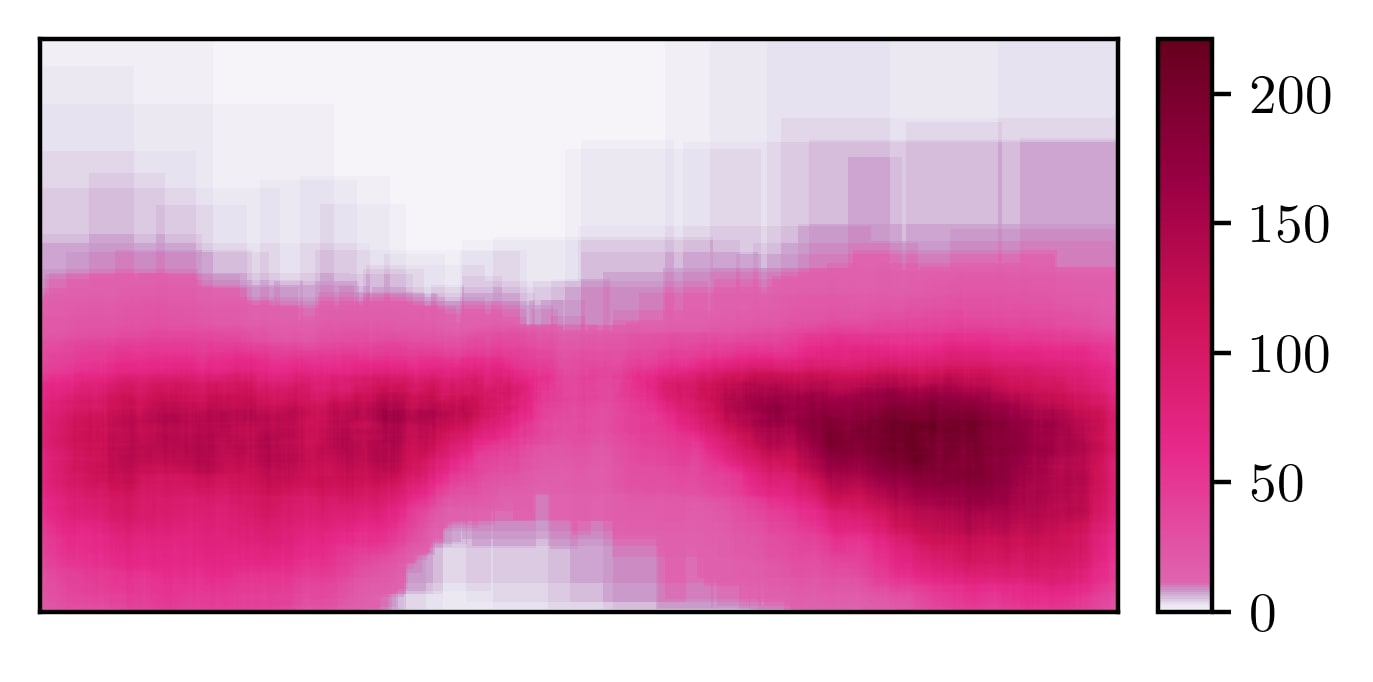}}\hfill
    \subfloat[CODA2022-ONCE]{\includegraphics[height=2.0cm]{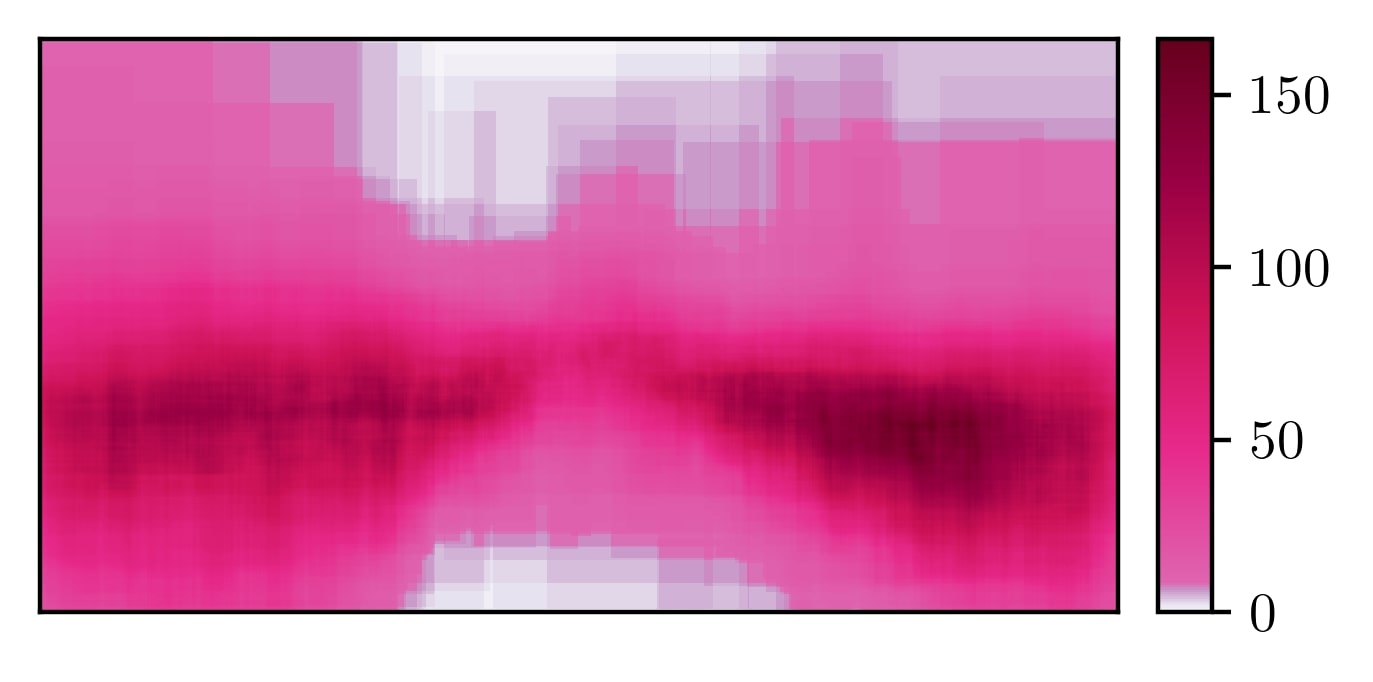}}\hfill
    \subfloat[CODA2022-SODA10M]{\includegraphics[height=2.0cm]{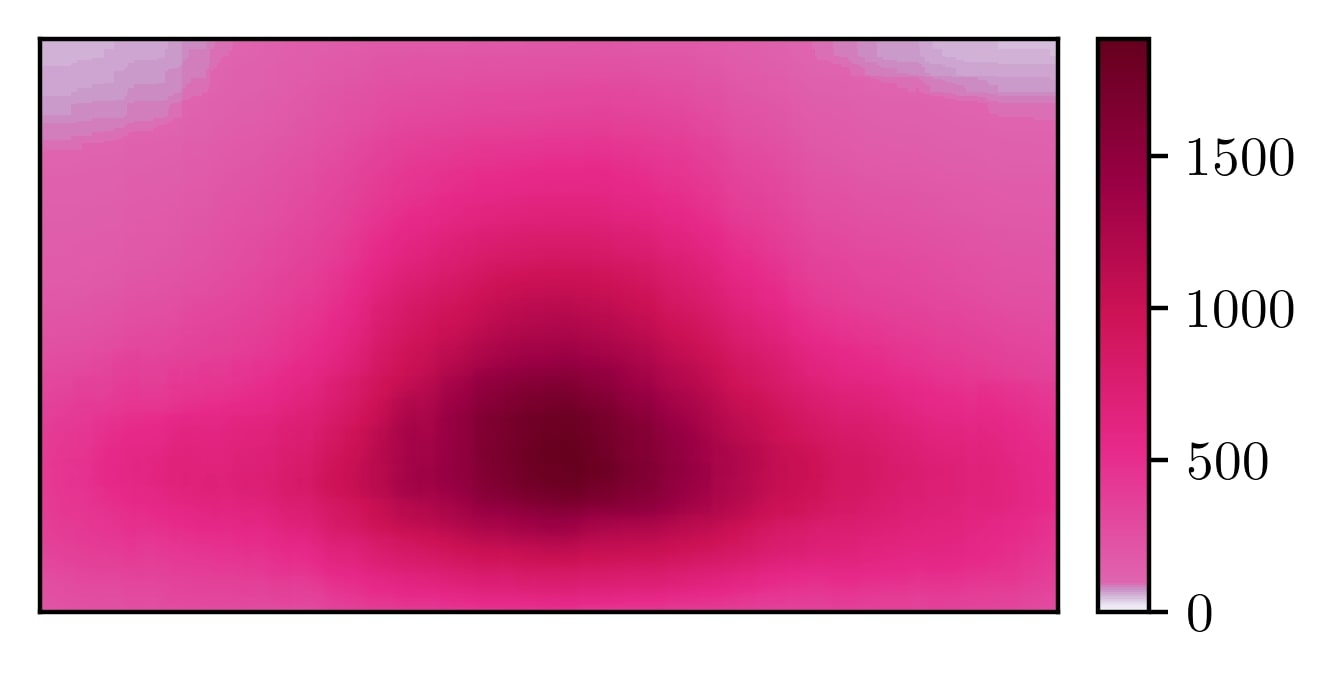}}\hfill
    \subfloat[Street Obstacle Sequences]{\includegraphics[height=2.0cm]{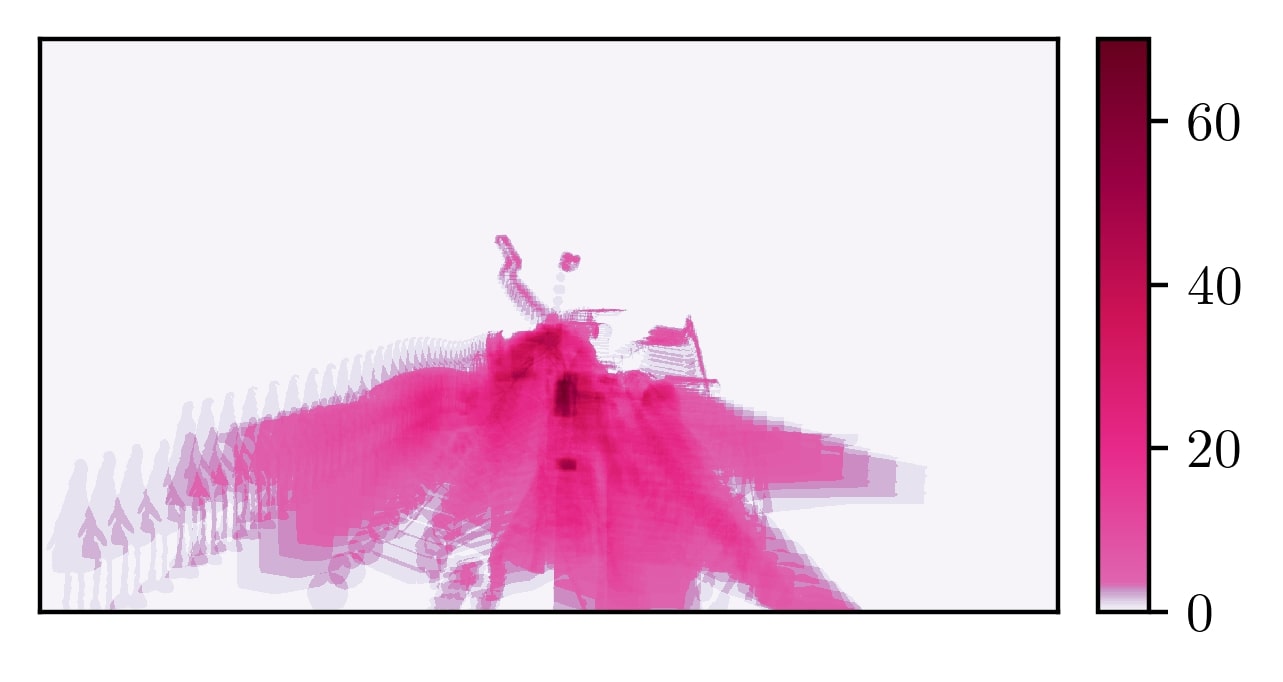}}\hfill
    \subfloat[CARLA-WildLife]{\includegraphics[height=2.0cm]{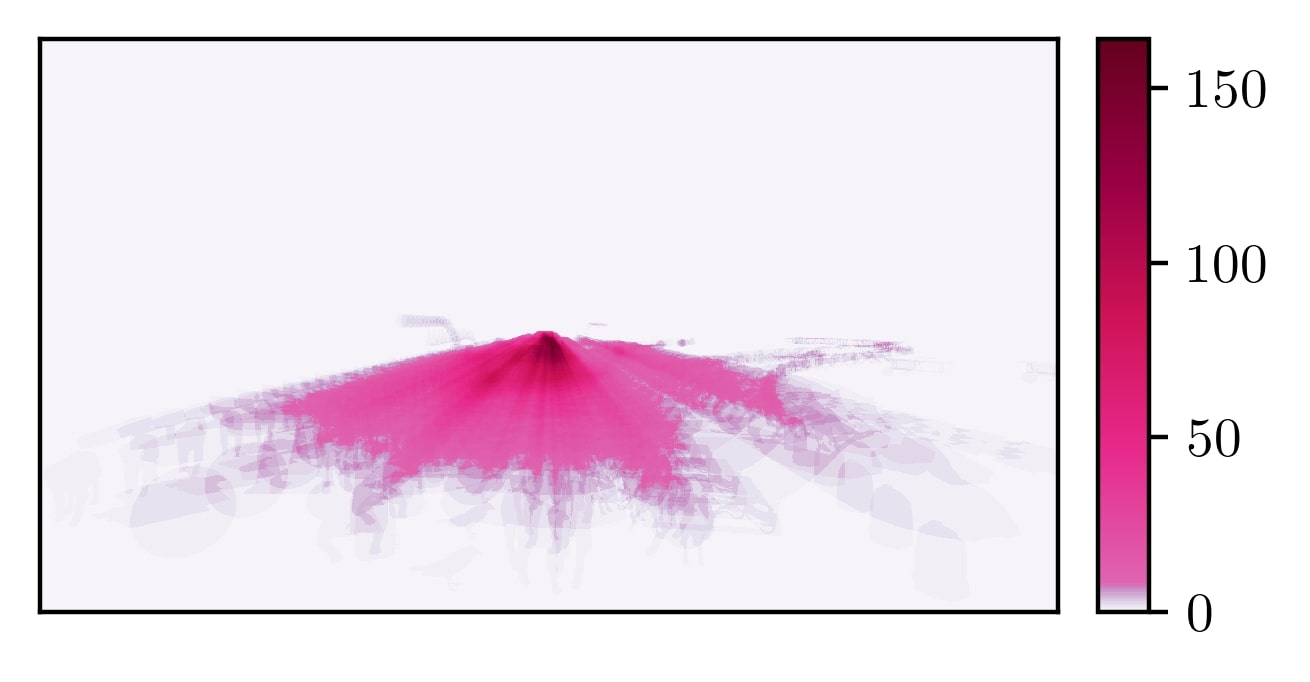}}
    
    \caption{Cumulated masks of all contained anomalies within the respective datasets.}
    \label{fig:anomalymasks}
\end{figure*}

\textbf{Realism.} Especially for anomalies which were generated by \textit{Simulation} or \textit{Data Augmentation}, the level of realism can vary strongly, as visible in Figure~\ref{fig:anomalymasks}. For example, anomalies in WD-Pascal and StreetHazards are often placed in implausible locations or scaled in unrealistic ways. However, such instances could appear in the real world, e.g., on billboards.

\textbf{Sensor Data.} Perception datasets with anomalies mostly provide camera data. The CODA Base datasets are the only exception and also include lidar point clouds; however, no anomaly labels are provided in 3D space. Crowdsourcing approaches make it hard to transfer detection methods to the perception system of an autonomous vehicle, as viewpoints vary strongly~\cite{abstraction}. Finally, representations of anomalies in camera data are not actionable for an autonomous driving system, however, this is a general computer vision issue.

\textbf{Regular Tasks.} Often datasets only provide labels for the anomalies. In this case only the task of anomaly detection is possible. However, it is generally of interest to be able to detect anomalies while still performing well at detecting or segmenting known classes.

\textbf{Domain Shift.} 
The recording of anomalies in the real world is time-consuming, as they are rarely present in ordinary street scenes and thus, have to be selected and placed manually. Furthermore, anomalies that would lead to dangerous driving situations cannot be captured. Consequently, anomaly techniques such as data augmentation and simulation emerged to tackle these issues. Simulation has the advantage of having full control. Thus, anomalies certainly do not appear in the training data. However, a natural domain gap to reality exists, so anomaly detection methods that perform well on synthetic data are not implicitly reliable on real-world data. The same holds for data augmentation, which mixes two domains, leading to unrealistic results. To ensure that methods really detect the anomalous objects that are pasted into the images, Fishyscapes pursues two strategies: Augmenting the Cityscapes images and pasting in objects from known classes. The first strategy prevents a method from only detecting pixels that differ from the non-augmented image, the second indicates whether only the domain shift is identified.

\textbf{Size.}
Datasets which include real-world scenes showing anomalies with respect to Cityscapes, i.e., which are recorded or collected from web resources, are usually very small. Lost and Found as the largest of these datasets only provides coarse annotations, followed by the Street Obstacle Sequences dataset, which however is highly redundant as the frames are extracted from $20$ video sequences. These datasets are mainly for evaluation purposes and not for training. They also provide a wide variety of different anomaly types, which is beneficial for evaluating anomaly detection but is a hindrance for further processing of these anomalies, e.g., in terms of image retrieval, clustering and incremental learning. While Vistas-NP and CODA are comparably large datasets, they are still not comparable to regular perception datasets with hundreds of thousands of frames.

\textbf{Similarity.}
Generating larger datasets as a combination of similar datasets requires that those have a proper anomaly technique, definition of normality and labeling policy. Such datasets include FS Lost and Found, RoadAnomaly21, RoadObstacle21 and SOS for semantic segmentation. In particular, it is not possible to combine other datasets in a meaningful way due to different labeling policies.

With this overview of datasets in the field of anomaly detection and mentioned challenges, we hope to contribute to larger, more diverse, or more specialized datasets in the future.
\section*{Acknowledgment}
\label{sec:acknowledgment}

This work results from the projects AI Data Tooling (19A20001J, 19A20001E) and AI Delta Learning (19A19013Q), funded by the German Federal Ministry for Economic Affairs and Climate Action (BMWK).

% -------------------------- REFERENCES -------------------------------

\hypersetup{
    colorlinks=true,
    citecolor=wong-green,
    linkcolor=wong-darkblue,
    filecolor=wong-pink,      
    urlcolor=wong-black,
    pdfpagemode=FullScreen,
    }

{\small
\bibliographystyle{IEEEtran}
\bibliography{references}
}

\end{document}